\documentclass[times,twocolumn,final]{elsarticle}

\usepackage{framed,multirow}

\usepackage{amssymb}
\usepackage{latexsym}
\usepackage{comment}
\usepackage{amsmath}
\usepackage{soul} 
\usepackage{caption}
\usepackage{mathtools}
\usepackage{booktabs}
\usepackage{multirow}
\usepackage{csquotes}

\usepackage{url}
\usepackage{xcolor}
\definecolor{newcolor}{rgb}{.8,.349,.1}

\newcommand{\red}[1]{{\color{red}#1}}

\newcommand{\TODO}[1]{\textbf{\color{red}[TODO: #1]}}

\usepackage{rotating} 
\usepackage{hyperref}

\usepackage[switch,pagewise]{lineno} 

\usepackage{multirow}
\usepackage{subcaption}

\usepackage[english]{babel}
\usepackage{amsthm}
\newcommand{\etal}{\textit{et al.}}

\theoremstyle{definition}

\usepackage{tikz}
\usetikzlibrary{positioning,arrows.meta,fit,backgrounds}

\definecolor{pcblue}{RGB}{52,115,183}
\definecolor{pclightblue}{RGB}{214,230,245}
\definecolor{pclightorange}{RGB}{255,231,204}
\definecolor{pclightgreen}{RGB}{214,240,214}
\definecolor{pclightpurple}{RGB}{232,222,248}

\tikzset{
  pipeline/.style={
    draw, rounded corners, thick,
    fill=pclightblue,
    minimum width=4.6cm,
    minimum height=1.2cm,
    align=center,
    font=\small
  },
  pipelinePH/.style={
    pipeline,
    fill=pclightpurple
  },
  modelblock/.style={
    pipeline,
    fill=pclightgreen
  },
  ipblock/.style={
    draw,
    rounded corners,
    thick,
    fill=pclightorange,
    minimum width=4.1cm,
    minimum height=1.3cm,
    align=left,
    font=\scriptsize
  },
  thickarrow/.style={->, thick, >=Latex}
}

\begin{document}


\begin{frontmatter}


\title{A Persistent Homology Design Space for 3D Point Cloud Deep Learning}

%


\author[smu]{Prachi Kudeshia}
\ead{prachi.kudeshia@smu.ca}

\author[smu]{Jiju Poovvancheri\corref{cor1}}
\ead{jiju.poovvancheri@smu.ca}

\cortext[cor1]{Corresponding author}

\author[smu]{Amr Ghoneim}
\ead{amrtsg@gmail.com}

\author[nfu]{Dong Chen}
\ead{chendong@njfu.edu.cn}

\address[smu]{Saint Mary's University, Halifax, Canada}
\address[nfu]{Nanjing Forestry University, China}



\begin{abstract}
Persistent Homology (PH) offers stable, multi-scale descriptors of intrinsic shape structure by capturing connected components, loops, and voids that persist across scales, providing invariants that complement purely geometric representations of 3D data. Yet, despite strong theoretical guarantees and increasing empirical adoption, its integration into deep learning for point clouds remains largely ad hoc and architecturally peripheral. In this work, we introduce a unified design space for Persistent-Homology driven learning in 3D point clouds (3DPHDL), formalizing the interplay between complex construction, filtration strategy, persistence representation, neural backbone, and prediction task. Beyond the canonical pipeline of diagram computation and vectorization, we identify six principled injection points through which topology can act as a structural inductive bias reshaping sampling, neighborhood graphs, optimization dynamics, self-supervision, output calibration, and even internal network regularization. We instantiate this framework through a controlled empirical study on ModelNet40 classification and ShapeNetPart segmentation, systematically augmenting representative backbones (PointNet, DGCNN, and Point Transformer) with persistence diagrams, images, and landscapes, and analyzing their impact on accuracy, robustness to noise and sampling variation, and computational scalability. Our results demonstrate consistent improvements in topology-sensitive discrimination and part consistency, while revealing meaningful trade-offs between representational expressiveness and combinatorial complexity. By viewing persistent homology not merely as an auxiliary feature but as a structured component within the learning pipeline, this work provides a systematic framework for incorporating topological reasoning into 3D point cloud learning.
\end{abstract}

\end{frontmatter}


\section{Introduction}
\begin{figure}[h]
\centering
\includegraphics[width=8cm]{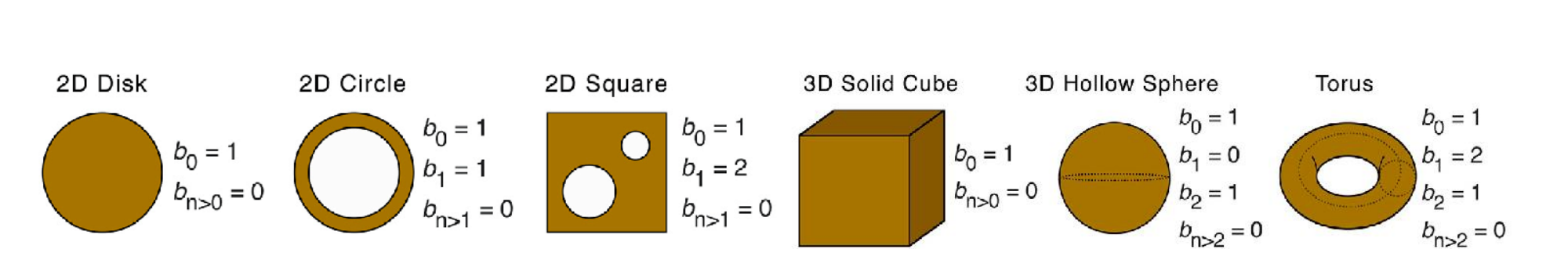}
\caption{Illustrative 3D objects with different topologies. Each object is labeled by its Betti numbers $(\beta_0,\beta_1,\beta_2)$: $\beta_0$ (0D holes = connected components), $\beta_1$ (1D holes = loops), and $\beta_2$ (2D holes = voids)~\cite{stenseke2021persistent}.}
\label{fig:betti-shapes}
\end{figure}

Topological Data Analysis (TDA) has emerged as a powerful mathematical framework for understanding the intrinsic structure of complex data beyond purely geometric or statistical representations~\cite{zia2024topological}. By focusing on properties that remain invariant under continuous deformations, TDA captures the shape of data in a way that complements conventional metric-based approaches. In the context of 3D shape and point cloud analysis, topology provides an abstract yet meaningful language for describing connectivity patterns, loops, and voids, which often underpin the semantics of objects and scenes. Unlike distance- or angle-based descriptors that are sensitive to sampling irregularities and local noise, topological invariants offer global and deformation-stable summaries of structure. For instance, the Betti numbers $(\beta_{0}, \beta_{1}, \beta_{2})$ (Fig. \ref{fig:betti-shapes}) succinctly quantify the number of connected components, tunnels, and cavities within a shape~\cite{stenseke2021persistent}, revealing information that remains consistent under smooth transformations. This invariance makes topological descriptors particularly valuable for analyzing geometric data acquired from real-world sensors, where occlusion, noise, and non-uniform density are prevalent.

Among the tools developed in TDA, Persistent Homology (PH) stands out as a principled method to extract and quantify topological features across multiple spatial scales. Starting from a discrete point set or mesh, PH builds a sequence of nested simplicial complexes by expanding local neighborhoods (e.g., via radius-based filtration) and tracks the birth and death of topological entities as the scale increases. The resulting persistence diagram (PD) or barcode encodes the lifespan of each feature, thereby revealing the hierarchical and multi-scale topology of the data. These representations have demonstrated robustness to noise and stability under perturbations~\cite{cohen2005stability,cohen2010lipschitz}, properties that are essential for real-world 3D applications. However, integrating PH into modern deep learning pipelines remains an open challenge. It involves non-trivial design choices, including the selection of appropriate filtrations and complexes, the transformation of PDs into differentiable vector representations~\cite{adams2017persistence}, and the maintenance of stability and expressiveness during learning. Bridging this gap between algebraic topology and neural representation learning promises not only to enrich the descriptive power of 3D models but also to endow them with a deeper understanding of the global geometric structure inherent in complex environments.

Recent advances have begun to address this challenge by embedding persistent homology directly into end-to-end trainable neural architectures. Early approaches incorporated handcrafted topological features as auxiliary descriptors for classification or retrieval, whereas more recent studies have introduced differentiable mechanisms that learn directly from PDs. Methods such as TopologyNet~\cite{zhou2022learning}, PersLay~\cite{carriere2020perslay}, and DeepSet-based persistence embeddings~\cite{hofer2017deep} exemplify this trend, enabling gradient-based optimization over inherently discrete topological structures. This emerging paradigm of topological deep learning enhances robustness, interpretability, and generalization while offering a principled bridge between local geometric encoding and global shape understanding. In particular, applications to 3D shape classification~\cite{ghosh2025taco}, semantic segmentation~\cite{liu2022toposeg}, and shape generation or synthesis~\cite{zhou2022learning} illustrate the breadth of topology-aware modeling in modern vision systems. These advances open a pathway for a principled exploration of how persistent homology can be embedded within neural architectures to achieve topology-aware learning in 3D shape analysis.

In this paper, we adapt and extend the growing body of work on persistent-homology-based deep learning (PHDL)~\cite{,zia2024topological,pun2018persistent} to the domain of 3D point cloud analysis. We use the term \textbf{3DPHDL} to refer to PH methods integrated into geometric deep learning pipelines for 3D point clouds. To structure this emerging methodology, we introduce a comprehensive \emph{design space diagram} (Fig.~\ref{fig:ph-designspace}) that captures the canonical PHML workflow outlining key components such as complex construction (e.g., Vietoris-Rips (VR), \v{C}ech, alpha, cubical), filtration design (distance-, function-, or learned-based), PH computation (barcodes or PDs), vectorization (persistence images, landscapes, and related embeddings~\cite{adams2017persistence,hofer2017deep}), neural architectures (convolutional neural network, graph neural network, transformer, autoencoder, simplicial neural network, MLP, etc.), and prediction (object classification, segmentation, reconstruction, regression, shape completion, etc.).  

Beyond this standard PHML workflow, our design space emphasizes that PH can play a far more integrated role in 3D deep learning architectures. Rather than functioning solely as an auxiliary descriptor appended to the network through vectorized diagrams or topological losses, PH can intervene at multiple stages of the pipeline and at multiple levels of abstraction. At the data level, PH can guide how point sets are sampled, partitioned, and represented, biasing the model toward regions with rich structural variation. Within the architecture itself, PH can inform neighborhood construction, message-passing patterns, and pooling strategies by identifying edges or regions that carry topologically salient information. During optimization, persistent features can modulate the learning dynamics, e.g., by shaping curricula, influencing sample selection, or determining when representations have stabilized sufficiently to permit early exits.

PH also offers a principled lens for designing self-supervised tasks and data augmentations, distinguishing between transformations that preserve essential homology and those that intentionally perturb it. At the output stage, PH can be used to assess the structural plausibility of predictions, guiding pseudo-label acceptance or providing topology-aware estimates of uncertainty. Finally, PH can act implicitly on the model itself, regularizing internal computations through mechanisms such as neural persistence or persistent Laplacians applied to feature graphs or weight structures. Taken together, these injection points reposition PH as a structural inductive bias that permeates the data, architecture, training process, and inference behavior, rather than a peripheral summary statistic. This expanded perspective motivates a broader exploration of 3DPHDL models, their theoretical properties, and their empirical impact on 3D classification and segmentation tasks.

We complement this conceptual design space with empirical demonstrations that instantiate different design choices across the stages of a 3DPHDL pipeline for point cloud classification and segmentation. Rather than evaluating each injection point individually, these experiments illustrate how representative combinations of complexes, filtrations, vectorizations, and integration strategies behave in practice. By comparing alternative configurations, we highlight how specific design decisions influence standard performance metrics such as accuracy, robustness to perturbations, and computational overhead, thereby providing practical guidance for constructing effective PH-augmented 3D learning systems.

\textbf{Our contributions are as follows:}
\begin{itemize}
    \item We develop a unified design space for 3DPHDL, highlighting the core PHML pipeline and six avenues for topological integration.
    \item We provide a detailed analysis of the theoretical and computational implications of integrating PH at various stages of the 3D learning process.
    \item We perform a controlled empirical evaluation of 3DPHDL, varying backbone models and PH vectorizations, and analyze their impact on accuracy, robustness, and computational efficiency for classification and segmentation.

\end{itemize}

The remainder of the paper is organized as follows:  Section~\ref{sec:math} reviews the mathematical background (simplicial complexes, homology, PH). Section~\ref{sec:lit} reviews the literature. Section~\ref{sec:ph-ml} describes the general PH-based learning pipeline and various ways to integrate PH into the PHDL pipeline. Section~\ref{sec:theorynperf} discusses key theoretical considerations (stability metrics, vectorization) and analyzes performance aspects (complex choices, algorithmic complexity, empirical robustness). Section~\ref{sec:experiments} presents empirical demonstrations of the 3DPHDL pipeline for two downstream tasks. Section~\ref{sec:conclusion} concludes and outlines future directions.
\begin{figure}[t]
\centering
\includegraphics[width=8 cm]{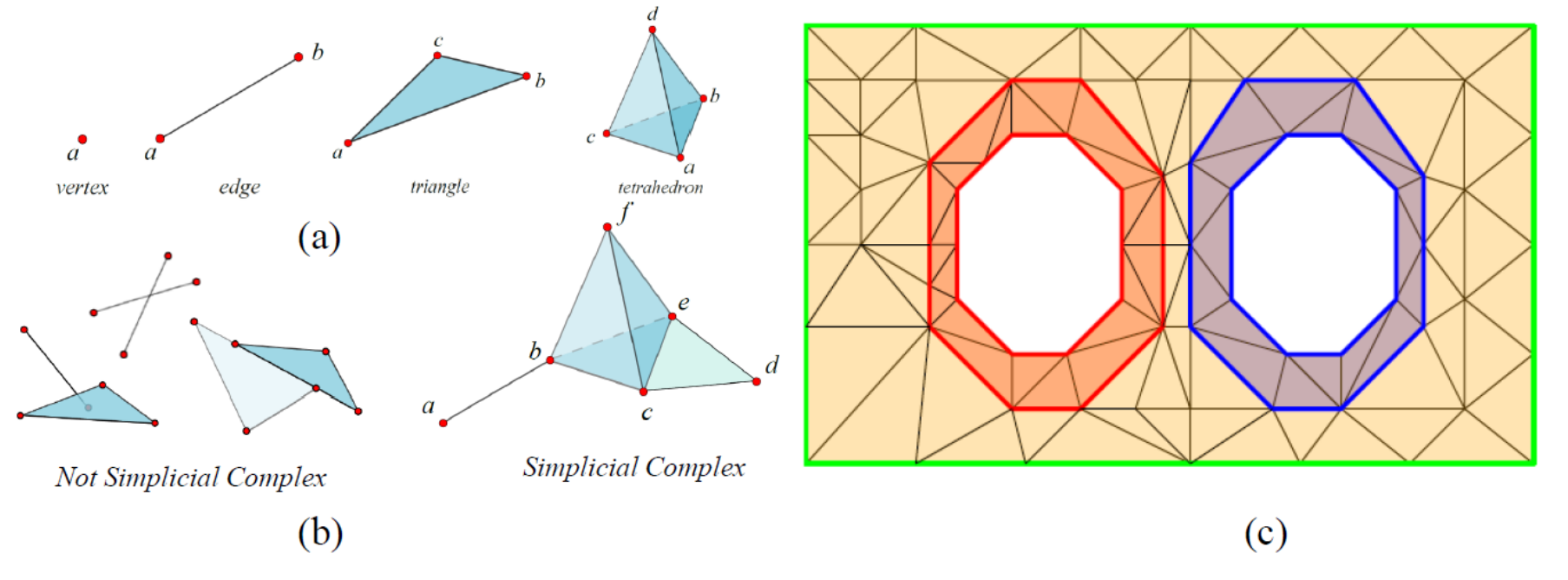}
\caption{(a) Example simplices: 0-simplex (point), 1-simplex (edge), 2-simplex (triangle), 3-simplex (tetrahedron). (b) A valid simplicial complex (right) vs. invalid configurations (left) where faces are missing~\cite{KovacevNikolic2012}. (c) Two individual 1-cycles (one blue and one red) exist in the complex; each cycle independently bounds a 2D chain of simplices~\cite{dey2022computational}}.
\label{fig:simplex}
\end{figure}

\section{Preliminaries}\label{sec:math}
In this section, we introduce the core topological constructs underlying PH and formalize the concept of PH with illustrations. Definitions and terminology are adopted primarily from Dey and Wang’s~\cite{dey2022computational} standard reference on computational topology analysis.

\paragraph{Simplicial Complexes} A finite 3D point set $X \subset \mathbb{R}^3$ is a discrete sample with no inherent topology. We impose the topology by building a simplicial complex on $X$. Formally, a simplicial complex $K$ on a vertex set $V$ is a collection of subsets (simplices) that is \emph{closed under taking faces}: if $\sigma \in K$ and $\tau \subseteq \sigma$ then $\tau \in K$. A $k$-simplex $\sigma$ is a set of $k+1$ vertices that are affinely independent (equivalently, the convex hull of $k+1$ points in general position). For example, a 0-simplex is a point, a 1-simplex an edge (pair of points), a 2-simplex a triangle (three vertices), and a 3-simplex a tetrahedron. Fig.~\ref{fig:simplex}(a) and (b) illustrate these simplex types and valid versus invalid simplicial complexes. The simplicial complex encodes the adjacency and connectivity of the point cloud in combinatorial form. 

\paragraph{Homology groups and Betti numbers} Once a simplicial complex $K$ is built, algebraic topology defines \emph{homology groups} $H_k(K)$ that capture $k$-dimensional holes in $K$. Concretely, one constructs the group of $k$-cycles $Z_k$ and the group of $k$-boundaries $B_k$ (collections of simplices that bound a $(k+1)$-chain), and forms the $k$-th homology group as the quotient
\[
H_k(K) = Z_k(K)/B_k(K),
\]
so that two cycles are identified if they differ by a boundary. Fig.~\ref{fig:simplex}(c) illustrates red and blue cycles that bound 2D chains of simplices, identifying them as boundaries. The \emph{Betti number} $\beta_k = \mathrm{rank}\,H_k(K)$ is the rank of $H_k$ and counts the number of independent $k$-dimensional holes. Equivalently, $\beta_0$ is the number of connected components, $\beta_1$ the number of independent loops or tunnels, and $\beta_2$ the number of voids (cavities). For instance, a sphere has Betti numbers $(\beta_0,\beta_1,\beta_2) = (1, 0, 1)$ (one connected component and one 2D void), while a torus has $(1,2,1)$. Computing $H_k(K)$ at a fixed scale reveals global shape features but can be sensitive to noise and choice of scale.
\begin{figure}[t]
\centering
\includegraphics[width=8 cm]{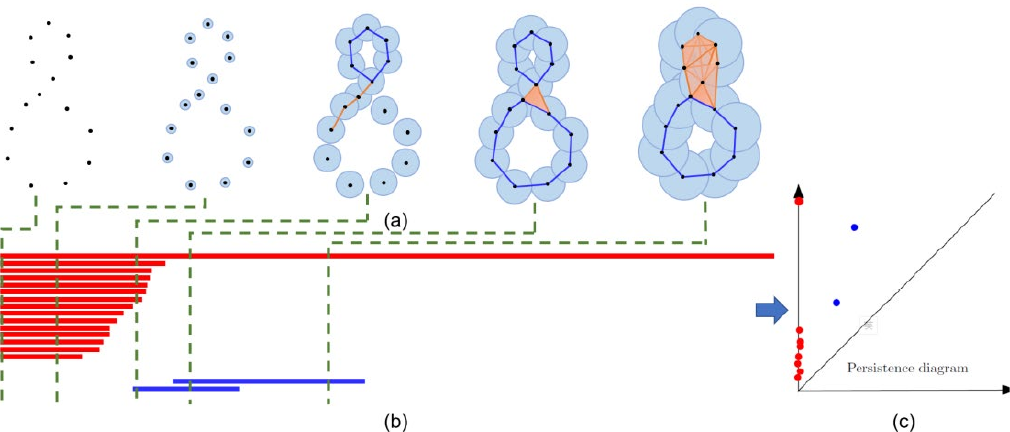}
\caption{Illustration of persistent homology computation. (a) A point cloud with growing ball radius $r$, showing emerging simplices (blue: existing, orange: new). (b) Resulting persistence barcode: each bar represents one topological feature (0D or 1D). (c) PD corresponding to (b), where each point $(b,d)$ represents a homological feature born at scale $b$ and dying at scale $d$~\cite{Wong2021PHGCN}.}
\label{fig:pers}
\end{figure}
\paragraph{Filtrations of Simplicial Complexes}
To obtain a multiscale view of topology, we build a \emph{filtration} of the complex by gradually adding simplices. A filtration of a simplicial complex $K$ is a nested sequence of subcomplexes
\[
\emptyset = K_0 \subseteq K_1 \subseteq \cdots \subseteq K_m = K,
\]
so that each $K_i$ includes all simplices of $K_{i-1}$. More generally, one may index by a parameter $r$ (often a real scale): we require $K_r \subseteq K_{r'}$ whenever $r \le r'$ and $\bigcup_r K_r = K$. In practice, the index $r$ may represent, for example, a radius or threshold. As $r$ increases, new simplices (edges, triangles, etc.) are added to the complex according to some rule, yielding a sequence of growing complexes.


\paragraph{Persistent Homology}
PH tracks the birth and death of topological features across a filtration. Formally, one computes the homology groups $H_k(K_r)$ at each scale $r$, and considers the maps induced by inclusion $K_r \hookrightarrow K_{r'}$ (for $r \le r'$). As shown in Fig. \ref{fig:pers}(a), a $k$-dimensional feature (cycle) appears (\emph{birth}) at some parameter $b$ when it first lies in the homology of $K_b$, and disappears (\emph{death}) at parameter $d$ when it becomes a boundary in $K_d$. PH records each such feature as an interval $[b,d)$ representing its lifetime. Equivalently, each feature can be represented by a point $(b,d)$ in the \emph{persistence diagram}, where the first coordinate is the birth scale and the second is the death scale. The collection of all such points (one for each homology class) is the persistence diagram (PD) (Fig. \ref{fig:pers}(c)) or a ``barcode'' if drawn as horizontal bars (Fig. \ref{fig:pers}(b)). Features that persist over a long interval (points far from the diagonal in the diagram) are typically regarded as significant, whereas short-lived features are often attributed to noise. Thus, PH provides a concise multiscale summary of the topology of the data. 

Each of these concepts, i.e., simplicial complexes, filtration, and PH, is a standard tool in TDA, enabling rigorous multiscale study of the “shape” of data.

\section{PHDL in 3D Point Cloud Analysis}\label{sec:lit}

PH has found a growing role in 3D point cloud and shape analysis. Early works used PH as a hand-crafted feature for shape classification or retrieval. For example, PH summaries of point clouds can capture intrinsic shape properties and improve accuracy on benchmark datasets~\cite{wu20153d,Chang2015ShapeNet}. Barnes \emph{et al.} provided a comparative study of PH featurization methods and their performance on various datasets, highlighting that persistence landscapes, images, and kernel methods can be effective shape descriptors~\cite{barnes2021comparative}. 

\paragraph{3D Shape Classification} Several recent works have integrated PH directly into end-to-end networks. Zhou \emph{et al.} proposed \textit{TopologyNet}~\cite{zhou2022learning}, an architecture that learns to predict persistence features from raw point clouds, using them to improve an autoencoder and a GAN for point cloud classification and generation. This approach significantly reduced the computational time required to obtain topological features. Nishikawa \emph{et al.} introduced a learnable filtration for PH, showing that adaptively choosing the distance threshold can improve the classification accuracy on 3D object datasets~\cite{nishikawa2023adaptive}. Similarly, de Surrel \emph{et al.} introduced \textit{RipsNet}, a neural architecture that bypasses explicit PD computation by directly learning a vector representation of PDs from point clouds. RipsNet proved robust to noise and enabled efficient classification based on topological features~\cite{de2022ripsnet}. \textit{TACO-NET} \cite{ghosh2025taco} presented a novel 3D object classification framework that transforms a point cloud into a voxelized binary 3D image to extract distinguishing topological features and trains a lightweight 1D CNN on extracted topological features. Another recent work, \textit{TopoLayer} \cite{guan2025topolayer}, introduced a universal neural network layer that extracts global topological features from point clouds using PH. This work proposed two vectorization approaches to capture topological information. The integration of these vectorization techniques and topological layers significantly boosted the accuracy of established architectures such as PointMLP and PointNet++ on classification and segmentation benchmarks without requiring architectural redesigns.

\paragraph{Point Cloud Segmentation} Topology-aware segmentation methods incorporate PH losses to enforce the correct topology of predicted shapes. In an early work on PH-based point cloud segmentation, Beksi \etal~\cite{beksi20163d} computed the zeroth homology group of the VR complex for scene segmentation. This work introduced the utilization of topological persistence for point cloud processing. Liu \emph{et al.} (2022) proposed \textit{TopoSeg}, a point cloud segmentation module that added a PH-based loss to standard networks. By comparing the PH of the predicted labels with that of the ground truth, TopoSeg reduced topological errors (e.g., \ spurious or missing loops) and improved baseline segmentation accuracy~\cite{liu2022toposeg}. Wong and Vong ~\cite{Wong2021PHGCN} similarly incorporated PH into a graph convolution network \textit{PHGCN} for fine-grained 3D segmentation, using a PD Loss to capture multi-scale structure. Their experiments demonstrated significant improvements in complex part segmentation tasks.

\paragraph{Shape Generation and Completion} PH has also been used to guide 3D shape generation and completion. Zhou \emph{et al.} (2022) used a TopologyNet branch within a GAN to enforce that the generated point clouds share the topological signatures of training shapes~\cite{zhou2022learning}. More recently, Guan \emph{et al.} (2025) introduced \textit{TopoDiT-3D}, a diffusion transformer that integrates persistence images as global tokens. TopoDiT-3D builds a Vietoris–Rips filtration of each shape and extracts persistence images of connected components, loops, and voids, incorporating this multi-scale topology into the generative model~\cite{guan2025topodit}. In another point cloud completion work, Pathak \emph{et al.} (2025) argued that real-world incomplete clouds exhibit rich 0D/1D topology, and show that including 0D homology (a skeletal prior of the complete shape) helps the network produce topologically-consistent completions~\cite{pathak2025revisiting}. Similarly, a recent generative framework TopoGen \cite{hu2025topogen}, bridged the gap between local geometry and global topology by utilizing PH in terms of Betti numbers and persistence points as conditional guidance within a latent diffusion model, which enables the synthesis of 3D shapes that are both geometrically diverse and topologically controllable from sparse inputs or sketches.

\paragraph{Pose Estimation and Alignment}
PH facilitates direct 3D alignment and 6D pose estimation by identifying features that remain stable across different viewpoints. Dey et al. (2010)~\cite{dey2010persistent} laid the groundwork by demonstrating that the birth-death values of topological features provide a more reliable signal for aligning complex 3D shapes than noise-sensitive geometric coordinates. A foundational work by Bonis et al. (2016) \cite{bonis2016persistence} further proved that persistence-based pooling can extract prominent geometric features (e.g., human extremities) into fixed-size vectors. Although their focus was pose classification, they demonstrated that topological persistence diagrams serve as stable, rotation-invariant descriptors for non-rigid shapes. Building on these principles, the \textit{TG-Pose} framework \cite{zhan2024tg} recently introduced a differentiable PH-based topological feature predictor. By extracting multi-scale structural signatures, TG-Pose effectively guides the estimation of an object’s rotation, translation, and scale, achieving robustness in scenarios where traditional geometric-only methods often fail due to occlusion.

\begin{figure*}[!th]
\centering
\includegraphics[width=16 cm]{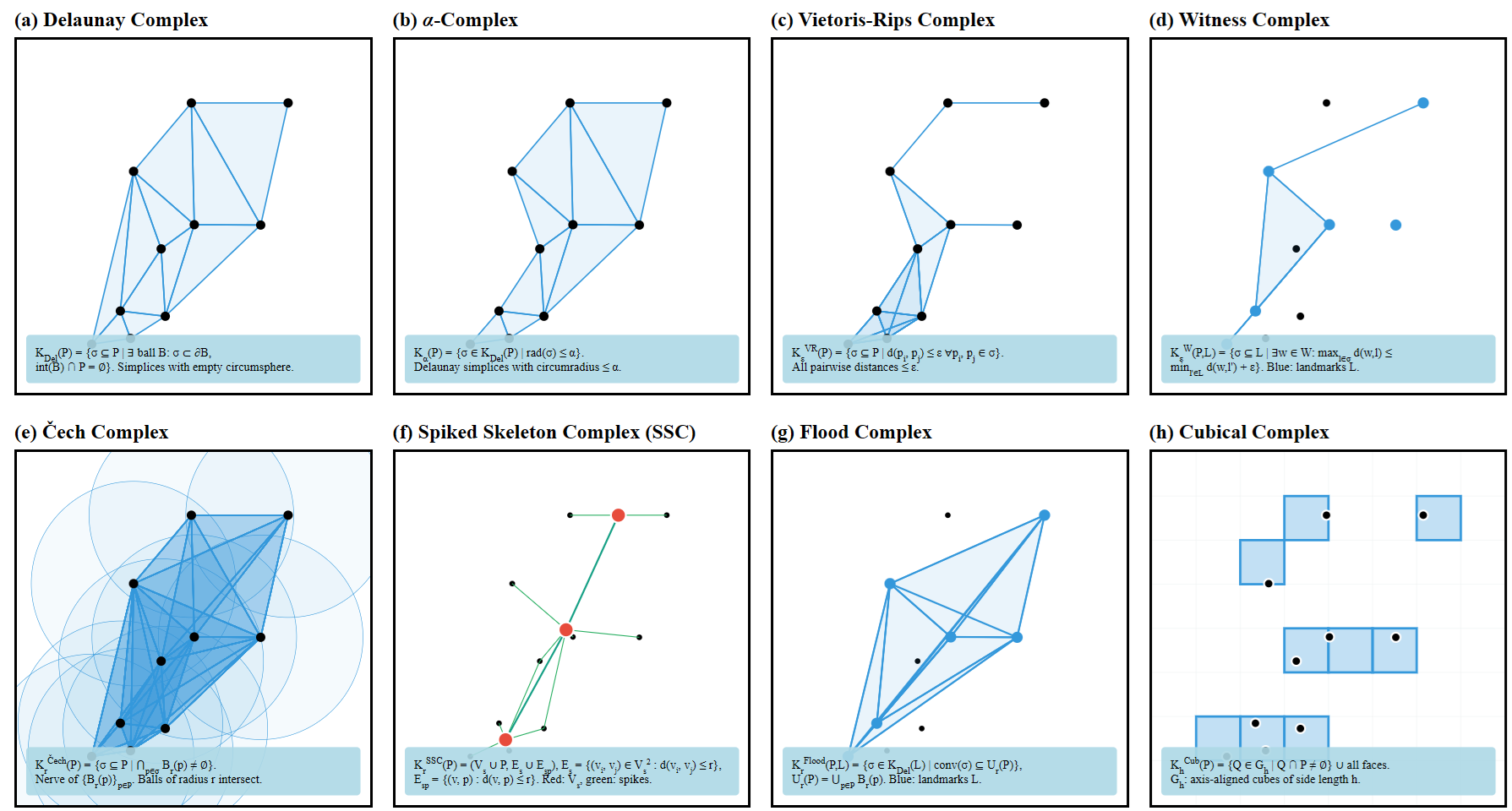}
\caption{
Representative constructions along the \emph{Complex Construction} axis of the 3DPHDL design space.
Let $P=\{p_0,\dots,p_n\}\subset\mathbb{R}^d$ and denote by $\mathcal{K}_\theta(P)$
a simplicial (or cubical) complex built on $P$ under construction parameter $\theta$
(e.g., scale, landmark subset, or grid resolution).
We use $d(\cdot,\cdot)$ for the Euclidean distance,
$B_r(x)=\{y\in\mathbb{R}^d \mid d(x,y)\le r\}$ for the closed ball of radius $r$ centered at $x$,
$\mathrm{conv}(\sigma)$ for the convex hull of $\sigma\subseteq P$,
$\mathrm{rad}(\sigma)$ for the circumradius of simplex $\sigma$,
$L\subseteq P$ for a landmark subset,
and $W=P\setminus L$ for the witness set.
(a) \textbf{Delaunay complex} \cite{delaunay1934sphere} $\mathcal{K}_{\mathrm{Del}}(P)$: simplices whose circumspheres are empty of other points in $P$.
(b) \textbf{$\alpha$-complex} \cite{edelsbrunner2003shape} $\mathcal{K}_{\alpha}(P)$: subcomplex of the Delaunay triangulation containing simplices with $\mathrm{rad}(\sigma)\le \alpha$.
(c) \textbf{Vietoris-Rips complex} \cite{gromov1987hyperbolic, vietoris1927hoheren} $\mathcal{K}^{\mathrm{VR}}_{\varepsilon}(P)$: simplices included when $d(p_i,p_j)\le \varepsilon$ for all vertex pairs in $\sigma$.
(d) \textbf{Witness complex} \cite{de2004topological} $\mathcal{K}^{\mathrm{W}}_{\varepsilon}(P,L)$: simplices $\sigma\subseteq L$ supported by witnesses in $W$ under proximity constraints.
(e) \textbf{\v{C}ech complex} \cite{vcech1932theorie} $\mathcal{K}^{\mathrm{Cech}}_{r}(P)$: simplices whose balls $\{B_r(p)\}_{p\in\sigma}$ have non-empty common intersection.
(f) \textbf{Spiked Skeleton Complex (SSC)} \cite{kudeshia2024learning} $\mathcal{K}^{\mathrm{SSC}}_{r}(P)$: skeleton nodes connected to nearby points within radius $r$, where skelton nodes are constructed through PCA based clustering.
(g) \textbf{Flood complex} \cite{graf2025flood} $\mathcal{K}^{\mathrm{Flood}}_{r}(P,L)$: Delaunay simplices over landmarks retained if $\mathrm{conv}(\sigma)\subseteq \bigcup_{p\in P} B_r(p)$.
(h) \textbf{Cubical complex} \cite{kaczynski2004computing} $\mathcal{K}^{\mathrm{Cub}}_{h}(P)$: grid-aligned cubes of side length $h$ intersecting $P$, together with their faces.
All constructions are illustrated in $\mathbb{R}^2$ for visualization clarity but admit natural generalizations to $\mathbb{R}^d$.
}
\label{fig:complexes}
\end{figure*}
\section{The Design Space for 3DPHDL}\label{sec:ph-ml}
\begin{figure*}[!th]
\centering
\includegraphics[width=16 cm]{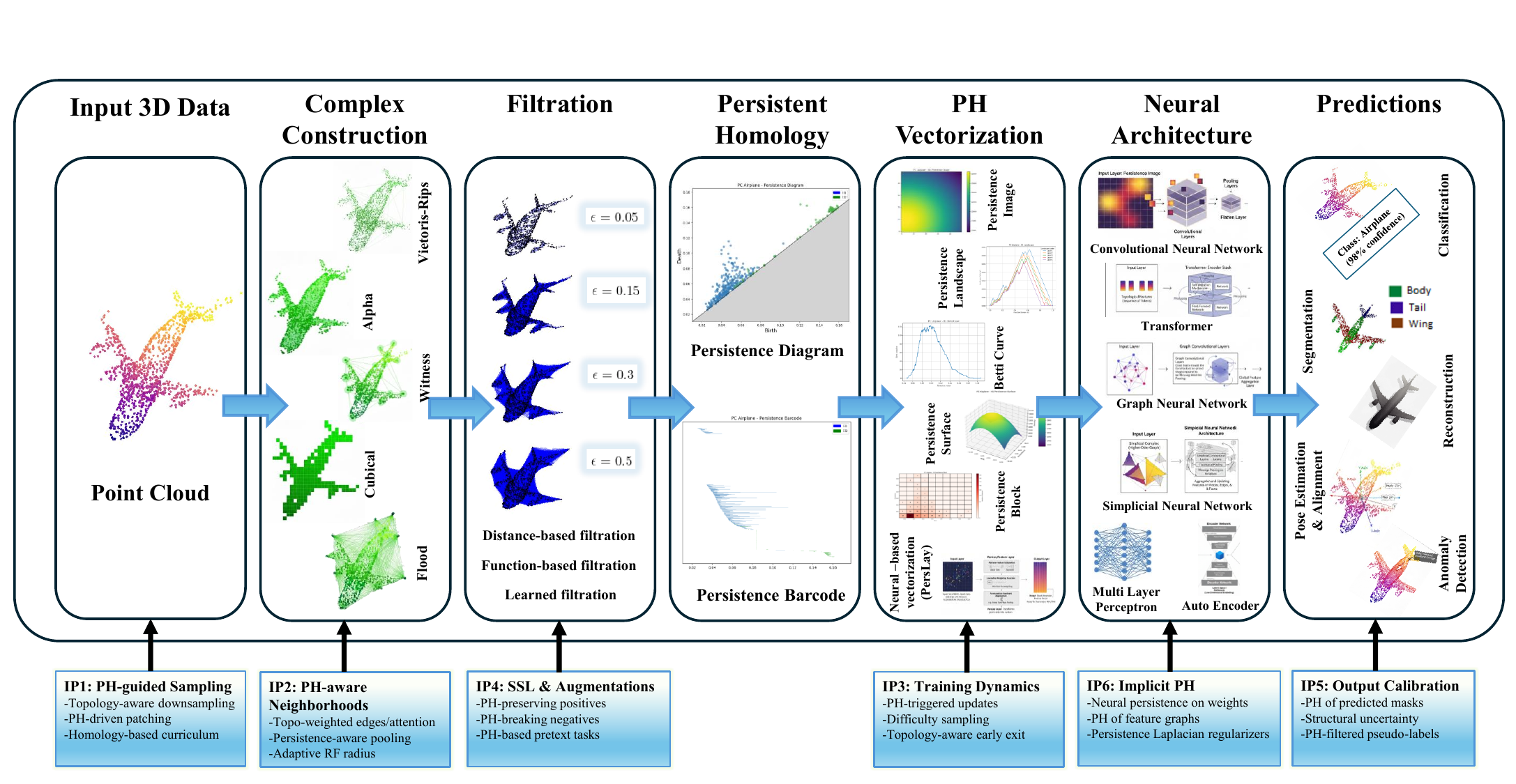}
\caption{
A design-space visualization for integrating persistent homology into 3D point-cloud learning pipelines. The horizontal band illustrates the standard PHML workflow
(raw data $\rightarrow$ complex $\rightarrow$ filtration $\rightarrow$ PH $\rightarrow$ vectorization $\rightarrow$ neural architectures $\rightarrow$ predictions).
Six injection points (IP1--IP6) appear as lateral modules, indicating where PH can reshape sampling (IP1), neighborhood/message passing (IP2), self-supervision/augmentations (IP4), training dynamics (IP3), implicit architectural regularization (IP6),
and output calibration (IP5).}
\label{fig:ph-designspace}
\end{figure*}

Fig.~\ref{fig:ph-designspace} illustrates the general design space of 3DPHDL models. We consider point clouds as input, as our focus is on neural architectures designed for point clouds. We organize the design space into two components: the overall pipeline and the injection points for PH.

\subsection{3DPHDL: A General Pipeline}
Most existing PHDL architectures for 3D shapes follow a relatively standard pattern: construct a simplicial or cubical complex, run a filtration, compute PDs, vectorize them (e.g.~persistence images or landscapes), and fuse the resulting descriptors with geometric features via concatenation and/or a topology-aware loss~\cite{carriere2020perslay,Wong2021PHGCN,gabrielsson20topology, hensel21survey}.

\paragraph{Complexes} A key design choice is the type of complex used to encode topology: popular options include the \v{C}ech \cite{vcech1932theorie} or VR \cite{gromov1987hyperbolic} complex (based on ball intersections) and the alpha complex \cite{edelsbrunner2003shape} (based on Delaunay triangulation \cite{delaunay1934sphere} for point clouds), or cubical complexes for voxel grids~\cite{gutierrez2012persistent, hu2024topology}. For example, VR and alpha complexes have been widely used for point clouds in PH-GCN~\cite{Wong2021PHGCN} and other recent works~\cite{zhou2022learning,nishikawa2023adaptive,de2022ripsnet} due to their computational tractability, whereas implicit or volumetric shapes often use cubical complexes~\cite{gutierrez2012persistent,hu2024topology}. Fig. \ref{fig:complexes} shows a representative set of existing simplicial complex constructions available for the Complex Construction axis of the 3DPHDL design space.

Though VR complexes \cite{gromov1987hyperbolic} are predominantly adopted due to implementation simplicity and GPU compatibility, Delaunay-based constructions \cite{delaunay1934sphere} provide a geometry-adaptive alternative that mitigates simplicial redundancy and may yield more stable topological signatures in non-uniformly sampled 3D point clouds. Despite their theoretical advantages and foundational role in alpha complexes \cite{edelsbrunner2003shape}, their explicit integration into modern PH-based deep learning architectures remains underexplored. Based on the concept of skeletal abstraction, a recently proposed compact representation, the spiked skeleton complex (SSC), captures the core skeletal structure of 3D point clouds \cite{kudeshia2024learning}. By leveraging a PCA-based construction complemented with spatial locality information, SSC serves as a global context descriptor that enhances robustness and discriminative power in 3D object classification architectures in a dual-stream setting.

On the other hand, witness complexes \cite{gromov1987hyperbolic} are considered an approximation to restricted Delaunay triangulation that sidesteps the curse of dimensionality, making them a computationally efficient representation for large, high-dimensional point clouds. By utilizing a small set of \enquote{landmarks} and a larger set of \enquote{witnesses}, they capture the underlying topology of complex metric spaces that would otherwise be computationally infeasible. However, the primary bottleneck remains landmark selection, as the quality of the topological approximation is highly sensitive to how well the chosen subset represents the geometry of the underlying manifold \cite{de2004topological}.

Inspired by the principles of alpha and witness complex, a recently introduced representation, the Flood complex \cite{graf2025flood} is specifically designed to represent large point clouds and allows for efficient PH computation up to dimension 2 on point clouds comprising several millions of points. From a design-space perspective, the Flood complex introduces a promising yet underexplored axis in 3DPHDL: adaptive filtration construction. While most existing PH-augmented architectures rely on VR complexes due to implementation simplicity, their combinatorial growth limits scalability for dense point clouds. The Flood complex’s propagation-based construction could enable topology-aware feature learning at scale, particularly when coupled with hierarchical sampling strategies or sparse convolutional backbones. Investigating such integrations remains an open research direction in topological deep learning.
 
\paragraph{Filtration} 
In TDA, a \emph{filtration} is a parameterized sequence $\emptyset = K_0 \subseteq K_1 \subseteq \cdots \subseteq K_m = K$ of the final complex $K$ where with each subcomplex $K_i$, simplices are added one by one according to a rule (often a scale parameter $r$). During the filtration,  persistent structural signals are distinguished from the transient \enquote{noise} by monitoring how topological features such as connected components or holes appear and disappear across this sequence ~\cite{dey2022computational,otter17roadmap, edelsbrunner2010computational}. An indexing function $f: K \to \mathbb{R}$ is a function that determines the order in which simplices are added to the complex. Based on the indexing function, filtration is generally categorized into three categories: distance-based filtration, function-based filtration, and learned filtration.

Distance-based or geometric filtration, such as Vietoris-Rips (VR) filtration and alpha filtration, is the most common filtration used on metric point clouds, where inclusion of a simplex in the next subcomplex is determined by the Euclidean distance between its vertices. For example, in VR filtration, a simplex is added at scale $r$ if all its vertices are within distance $r$ of each other. Similarly, in alpha filtration, a simplex is included in the subcomplex based on the intersection of balls centered at points restricted to their Voronoi cells \cite{edelsbrunner2010computational}.

Unlike geometric filtration, which uses raw distance, function-based or sublevel set filtration utilizes a scalar function $f$ defined on the vertices (and extended to simplices) to dictate the sequence. These are subcategorized as density-based filtration and height filtration. Density-based filtration uses a density estimator like the Distance to Measure (DTM) to include simplices in high-density regions that appear first, while avoiding the topological noise in low-density regions that appear much later, and hence allows for more robust topological signatures ~\cite{chazal2011geometric}. On the other hand, height filtration follows a specific directional coordinate, i.e., the z-axis, for the selection of simplices for the next subcomplex, which is useful for analyzing the morphology of 3D objects or surfaces ~\cite{turner2014persistent}.

In contrast to distance-based and function-based filtration, which are manually defined by geometry or density, learned filtration uses neural networks to parameterize the filtration function $f$. This filtration treats the construction of the PD as a differentiable step, while the model updates its weights via backpropagation based on a loss function defined directly on the topological output. Consequently, the model learns an optimal filtration that highlights the specific topological features most relevant to a given task, such as graph classification or shape matching~\cite{carriere2020perslay, hofer2017deep}. This approach is increasingly pivotal in modern machine learning, as it allows topological signatures to be task-adaptive rather than fixed.

\paragraph{Persistent Homology} The birth and death of each homology class is recorded in a \emph{persistence diagram} (PD): a multiset of points $(b_i,d_i)$ in the plane, each encoding a feature born at scale $b_i$ and died at $d_i$ (see Fig.~\ref{fig:pers}). Equivalently, one can represent the same information as a \emph{barcode} of intervals $[b_i,d_i]$. The collection of all $k$-dimensional features is denoted $L^k = \{(b^k_i,d^k_i)\}$. In general, in reference to point cloud-based topological analysis, features that persist a long time (large $d_i-b_i$) often correspond to prominent shape traits, whereas short intervals may be noise.

PH is computed by linear algebra on the boundary operators of the complex~\cite{dey2022computational}. PD (or barcode) is a complete topological summary of the shape across scales~\cite{zia2024topological,dey2022computational}. In practice, one obtains PDs using PH libraries such as GUDHI~\cite{Bauer2014}, PHAT~\cite{Bauer2017}, R-TDA~\cite{Fasy2014}, Perseus~\cite{Nanda2013}, or JavaPlex~\cite{Tausz2011}. For large point clouds, it is common to use the VR complex up to $H_2$ (since ambient 3D does not have higher holes), or the alpha complex to limit computation \cite{gutierrez2012persistent,mishra2023stability}.

\paragraph{Feature Vectorization}
PDs are not directly amenable to most neural networks, which expect fixed-size vectors. A popular approach is to \emph{vectorize} the PD, e.g., discretize the birth-death plane into a grid and compute a \emph{persistence image} (PI)~\cite{adams2017persistence}. Here, each diagram point contributes a weighted Gaussian to a smooth “persistence surface” which is then integrated over pixels. PI is stable under perturbations (Lipschitz stability w.r.t. Wasserstein distance)~\cite{adams2017persistence}, making it a robust feature. Other representations include persistence landscapes (PL) or topological descriptors~\cite{hofer2017deep, zeppelzauer2018study}. A complementary approach is to use kernel methods on diagrams~\cite{hofer2017deep}, but existing 3D works ~\cite{zhou2022learning,Wong2021PHGCN,de2022ripsnet} mostly prefer vectorization for computational efficiency.

PL provides a functional summary of a PD that maps topological features to a sequence of piecewise linear functions \cite{bubenik2015statistical}. It encodes the prominence of features such that the $k$-th highest peak corresponds to the $k$-th most persistent generator. This representation is particularly advantageous because it embeds the non-linear space of PDs into a separable Banach space, allowing for seamless integration into various machine learning frameworks.

A simpler alternative to the high-dimensional surfaces of PI or the nested functions of PL is the Betti curve, which provides a one-dimensional summary of topological features \cite{hayakawa2022quantum}. The Betti curve effectively counts the number of independent topological features at each scale by tracking its value as a discrete or continuous function of the filtration parameter. However, it discards the specific birth-death correlations of individual features. Nevertheless, its high level of interpretability and low computational overhead make it a popular baseline for analyzing global structural transitions \cite{topaz2015topological, bai2024topological}.

In contrast to predefined vectorization such as PI or PL, Carrière et al., 2020 \cite{carriere2020perslay} use a versatile neural network layer, \textit{PersLay}, designed to learn task-specific representations of PDs. PersLay treats PD points as an unordered point set and applies a permutation-invariant architecture comprising a transformation function, a weight function, and a pooling operation to optimize the topological summary for a particular learning objective. This framework is highly flexible, as it can recover many existing vectorization schemes by simply fixing its internal parameters.

In contrast to the aforementioned vectorization schemes, which are primarily designed for single-parameter filtration, the Generalized Rank Invariant Landscape (GRIL) extends the vectorization to the multi-parameter setting \cite{xin2023gril, mukherjee2024d}. GRIL addresses the complexity of 2-parameter persistence modules and allows for the integration of richer, multi-dimensional topological data into standard machine learning architectures like Graph Neural Networks (GNNs) without the computational bottlenecks typically associated with multi-persistence.

\paragraph{Neural Architecture}
Beyond topological construction and vectorization, the choice of \emph{neural backbone} plays a central role in 3DPHDL. Existing architectures for 3D learning can be broadly categorized into point-based, graph-based, convolution-based, and transformer-based models. Point-based networks such as PointNet~\cite{pointnet} and PointNet++ \cite{qi2017pointnet++} directly consume unordered point sets using symmetric aggregation functions. Graph-based methods, including DGCNN \cite{wang2019dynamic}, dynamically construct neighborhood graphs to capture local geometric relationships. Convolutional approaches extend spatial convolutions to irregular domains via kernel point convolutions or sparse voxel convolutions~\cite{Thomas2019KPConv, choy20194d}. More recently, transformer-based models leverage self-attention mechanisms to model long-range dependencies in point clouds~\cite{zhao2021point,Guo2021PCT}.

Within a 3DPHDL framework, topological descriptors can be fused with intermediate geometric features at various stages, such as early fusion (input-level concatenation), mid-level feature integration, or late fusion before prediction~\cite{Wong2021PHGCN,de2022ripsnet}. Importantly, the neural architecture determines how effectively geometric and topological signals interact, influencing expressivity, scalability, and robustness. From a design-space perspective, the backbone architecture constitutes an independent axis, orthogonal to the choice of complex, filtration, and PH vectorization.

\paragraph{Predictions} The final stage of a 3DPHDL pipeline concerns the \emph{prediction space}, i.e., the type of output the network is trained to produce. In 3D learning tasks, predictions typically fall into five categories: \emph{classes}, \emph{segments}, \emph{scalars}, \emph{fields}, and \emph{synthesis}. For global \emph{classification}, the network outputs categorical logits representing object-level labels~\cite{wu20153d,pointnet}. For \emph{segmentation}, the model produces per-point or per-voxel categorical predictions~\cite{Wong2021PHGCN,qi2017pointnet++,Liu2022}. In \emph{regression} settings, the output consists of continuous scalar or vector-valued quantities such as geometric attributes or physical parameters~\cite{zhou2022learning}. More generally, modern 3D learning frameworks predict continuous spatial \emph{fields}, including occupancy functions and signed distance fields (SDFs), enabling implicit surface representations~\cite{mescheder2019occupancy,peng2020convolutional}. Finally, in \emph{shape synthesis} or generative modeling, the network produces novel 3D shapes in the form of point clouds, meshes, or implicit fields, typically using autoencoder, variational, or adversarial frameworks~\cite{achlioptas2018learning,park2019deepsdf}.


\subsection{Integration Points of PH in Point Cloud Architectures}
\label{subsec:ph-injection-points}

The standard 3DPHDL pipeline has proven effective, but treats PH as an \emph{auxiliary branch} bolted onto an otherwise conventional network. In this subsection, we argue that PH offers a richer design space. It can shape how point clouds are sampled, how neighborhoods and kernels are constructed, how the optimizer allocates its effort, and even how self-supervision and uncertainty estimation are defined. We outline six complementary ``injection points'' where PH can be woven into point cloud architectures without necessarily materializing as an explicit parallel head.

\subsubsection{PH-guided Sampling and Patching} 
The first locus is \emph{upstream}, before any backbone (PointNet~\cite{pointnet}, DGCNN~\cite{wang2019dynamic}, or PointTransformer~\cite{zhao2021point}) processes the data. Rather than viewing PH as a descriptor of an already-chosen representation, one can use it to decide \emph{what} the network should see and at what resolution. Conceptually, this leverages the robustness and scale-awareness of PH~\cite{otter17roadmap,edelsbrunner2010computational} to impose structure on common pre-processing steps such as downsampling and patching.
\begin{figure}[t]
\centering
\includegraphics[width=8 cm]{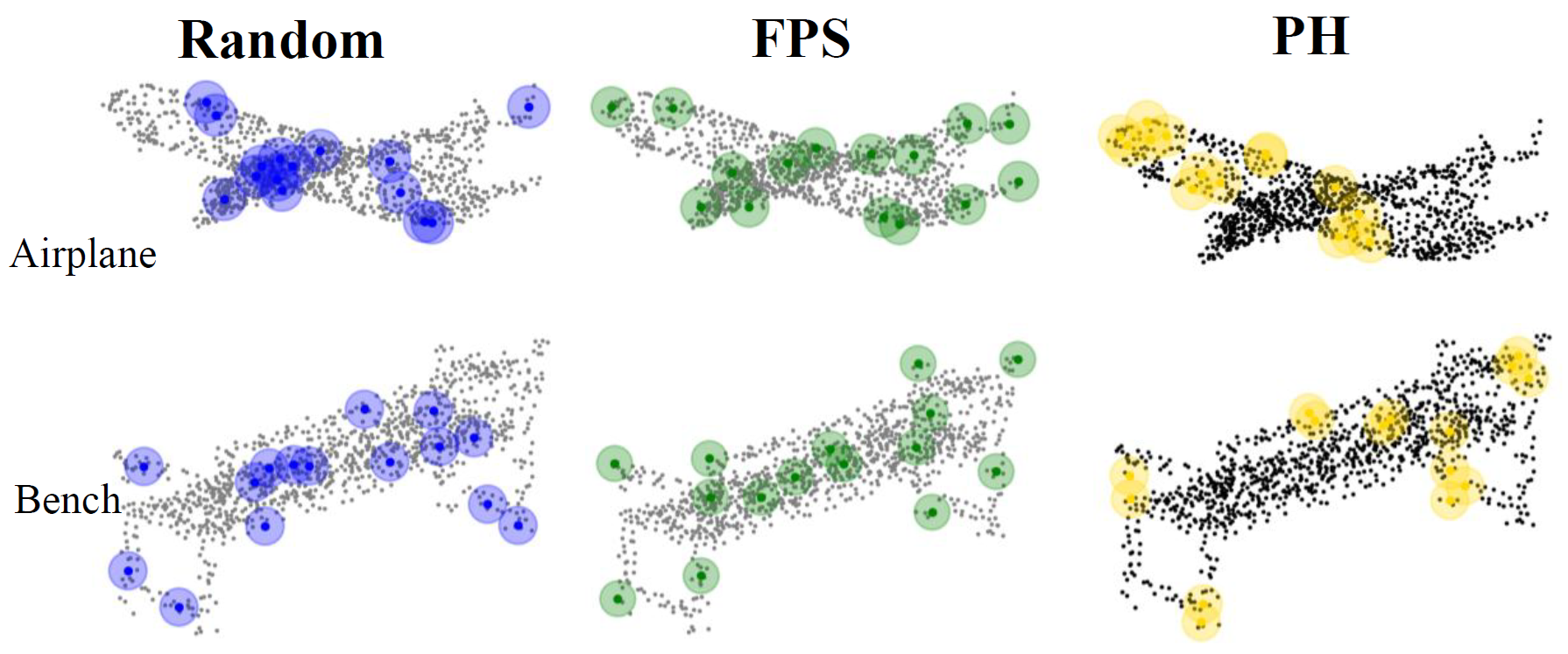}
\caption{Comparison of point cloud sampling strategies: Random, Farthest Point Sampling (FPS) \cite{moenning2003fast}, and Persistent Homology (PH)-guided sampling (left to right) with 15 samples per shape. PH-based sampling better preserves structurally and topologically salient features (e.g., boundaries and cycles), motivating its use over geometry-only approaches.}
\label{fig:samplings}
\end{figure}
\\
\indent A simple example is topology-aware downsampling. Given a point cloud, one computes a coarse PH signature on a reduced set of points or voxel centroids; for instance, $H_0$ and $H_1$ of a VR or alpha filtration, using libraries such as GUDHI~\cite{maria14gudhi} or Ripser~\cite{bauer2021ripser}. Regions with long-persistence components or cycles are marked as topologically "interesting", whereas flat, low-persistence regions are treated as "boring". Sampling then overshoots the former and aggressively thins the latter. In effect, PH becomes a principled attention mechanism on the raw geometry, biasing Farthest Point Sampling (FPS)/grid sampling toward structurally informative zones (thin beams, openings, façade articulations) and away from redundant planar patches. 

A closely related idea is PH-driven patch partitioning. Instead of defining local neighborhoods purely by Euclidean radius or $k$-nearest neighbors, one can cut along low-persistence valleys of an $H_0$ or $H_1$ landscape so that patches do not straddle topologically significant gaps or holes. Stage-1 tokens in a hierarchical encoder already respect key connectivity patterns. This is particularly appealing for fine-grained segmentation, where errors often occur near topological transitions (e.g. window embrasures vs. flat wall). 

PH-sampling identifies geometrically significant features, such as holes or voids, and selects them as representative seeds for local patches. By employing a persistence-based suppression logic, the algorithm filters out redundant candidates to ensure that the final seeds are both topologically prominent and spatially well-distributed. As illustrated in Fig. \ref{fig:samplings}, this approach offers a more structurally aware alternative to standard random sampling or FPS \cite{moenning2003fast}, presenting a promising direction for preserving critical shape semantics in point cloud processing.

Beyond acting as sampling and patching mechanisms, PH can also be used to organize training through a topology-aware curriculum over spatial resolution. The PD of each shape or scene provides a quantitative measure of homological complexity (e.g., count and total lifespan of long-persistence features in dimensions $k \in \{0,1,2\}$), which naturally distinguishes \emph{topologically simple} instances (few long-lived features) from \emph{topologically complex} ones (many long-lived cycles and cavities). Using this signal, one can schedule training such that simple shapes are introduced early using coarse PH-guided samples that retain only the most prominent large-scale structures, while increasingly complex instances are added later with progressively finer PH-based sampling that allocates more points to regions supporting long-persistence features. This curriculum does not alter the backbone architecture or loss; instead, it lives entirely in the data pipeline (which examples are seen when, and at what resolution), and thus echoes classical scale-space ideas in which structures are revealed and topological events tracked as the observation scale is gradually refined~\cite{lindeberg1994scale,romeny1997introduction}.


\subsubsection{PH-aware Neighborhood Graphs and Convolutions}

The second injection point is in the \emph{graph or neighborhood machinery} itself. Here, PH is no longer a side descriptor; it directly modulates how messages flow.

In graph-style point cloud networks (DGCNN~\cite{wang2019dynamic} or PointTransformer~\cite{zhao2021point}), one builds a $k$NN graph and then performs edge-based aggregation. PH can be used to assign a structural importance score to each edge; for instance, one can measure how contracting that edge changes the local $H_0/H_1$ within a small witness or VR complex around it. Edges whose contraction would destroy high-persistence cycles or split a persistent component are deemed topologically critical and receive higher weights or attention biases; edges in topologically redundant regions are downweighted. This is analogous in spirit to the way PHGCN injects persistence information into graph convolutions for 3D shape segmentation~\cite{Wong2021PHGCN}, but here the emphasis is on modifying the \emph{kernel geometry} rather than concatenating an external feature vector. 

A recent work \cite{kudeshia2024learning} extends the neighborhood machinery beyond simple pairwise connections through simplicial attention. Rather than restricting the attention mechanism to node-to-node interactions, this approach incorporates 0-simplices (node coordinates), 1-simplices (edge distances), and 2-simplices (triangle areas) of the alpha complex to compute attention coefficients that reflect the shared topological context of the neighborhood.  This allows the model to prioritize neighbors not just based on proximity, but on the higher-order spatial relationships of the underlying simplicial complex.

Pooling and unpooling can also become persistence-aware. During hierarchical encoding, features from points with high local persistence can dominate the aggregated representation, reflecting their role as carriers of global structure. Conversely, during upsampling for dense prediction, the decoder can be encouraged to reconstruct features near persistent structures (e.g.~structural edges, façade openings) more faithfully, akin to a topology-aware version of importance sampling. This perspective connects naturally to recent work on persistent Laplacians, where spectral quantities encode both homology and geometric connectivity~\cite{memoli22persistent_laplacians}.

A further refinement is to adapt the receptive-field radii based on local topological signals. In thin structures or around bottlenecks (where $H_1$ features are fragile), smaller radii maintain the separation of distinct components; in bulk regions with stable connectivity, larger radii are safe and computationally efficient. PH thus becomes a guide for spatial adaptivity, steering neighborhood size in a way that is difficult to achieve from purely metric heuristics.

\subsubsection{PH in Training Dynamics and Optimization}

PH can be integrated at the level of learning and optimization. 
The most direct mechanism is feature-level fusion: once topological descriptors (e.g., \ PIs, PLs, or diagram embeddings) are computed, they are concatenated with geometric features such as point coordinates, normals, or learned embeddings and fed into a classifier or segmentation backbone (DGCNN, PointNet++, transformers, etc.) \cite{zhou2022learning,Wong2021PHGCN,Liu2022}. 
This strategy enriches local geometric representations with global structural information, while preserving compatibility with standard architectures.

Beyond feature fusion, PH may directly shape the objective function through \emph{topological losses}. These losses measure a discrepancy between the predicted output’s PD and a ground-truth or reference PD, typically via Wasserstein or mean-squared-error distances \cite{zhou2022learning,Wong2021PHGCN,Liu2022}. For instance, PHGCN~\cite{Wong2021PHGCN} aggregates Wasserstein distances across part-level diagrams, while TopoSeg~\cite{Liu2022} minimizes MSE between predicted and target PIs. Such formulations explicitly regularize the connectivity and hole structures in the output space, encouraging topology-preserving predictions.

More globally, PH can influence the \emph{training dynamics themselves}, rather than serving solely as an auxiliary loss. Prior work on neural persistence~\cite{rieck2019neural} and topology of intermediate activations~\cite{hensel21survey} demonstrates that PH captures structural complexity within network representations. This motivates topology-aware optimization strategies that regulate \emph{when} and \emph{where} updates occur.

One possibility is PH-triggered update scheduling: computing a coarse PD (or scalar topological energy) on intermediate activations and tracking its change $\Delta E$ across epochs as a measure of representational stability. If topology stabilizes, certain layers may be temporarily frozen or computationally expensive modules skipped; if topology remains volatile, additional gradient steps or adaptive learning-rate adjustments may be allocated. In nested learning settings~\cite{behrouznested}, PH-derived signals could further gate inner versus outer loop updates.

Topology-based discrepancy measures also enable principled curriculum and hard-example mining~\cite{bengio2009curriculum}. Samples whose activation-level diagrams deviate significantly from class prototypes can be upweighted, prioritizing structural alignment over purely pointwise accuracy. Conversely, topologically aligned samples may be downsampled to mitigate overfitting.

Finally, PH offers a mechanism for topology-aware adaptive inference. If the PD of an intermediate layer matches a class-specific prototype within a predefined threshold, for example, calculated via Wasserstein or sliced-Wasserstein distance~\cite{carriere2017sliced}, the network may exit early. This enables adaptive depth, reducing inference cost for structurally simple inputs while retaining deeper processing where topology remains ambiguous.

\subsubsection{PH-guided Augmentations and Self-supervision}

The fourth integration point is the design of \emph{augmentations and self-supervised objectives}. PH offers a principled way to distinguish between transformations that should be treated as invariances versus those that should induce contrastive separation. This aligns with recent topological perspectives on contrastive and generative learning \cite{chen2024topogcl,chen2022topoattn,papamarkou2024position}, although these works primarily target graphs and abstract representations; here, the same principles are extended to 3D point clouds.

For positive pairs in contrastive or JEPA-style~\cite{saito2025point} training, one can encourage or select augmentations that preserve PH up to small perturbations of the diagrams: rigid motions, mild jitter, controlled density variations, or limited local deformations that leave PD essentially unchanged. Augmentations that significantly alter persistent cycles or components (e.g., removing entire beams, closing apertures) can be filtered out or downweighted. In this way, PH acts as a filter on the augmentation pipeline, enforcing that the model’s invariances remain consistent with the intrinsic shape topology.

Conversely, PH can be used to \emph{construct} strong negatives and out-of-context views by deliberately applying perturbations (cropping, hole punching, topological surgery) that induce clear changes in homology, and verifying via PH that the resulting PDs differ substantially from the original. These views become topologically distant contexts in JEPA or negative samples in InfoNCE~\cite{oord2018representation} objectives, sharpening the network’s sensitivity to structural changes. Graph-level work such as TopoGCL and TopoAttn-Nets already demonstrates the benefits of contrasting topological representations across augmentations~\cite{chen2024topogcl,chen2022topoattn}; the constructions proposed here transfer those principles to 3D point clouds by designing augmentations that explicitly modulate homology.

Finally, PH naturally yields pretext tasks such as predicting Betti curves, Euler characteristic curves, or coarse PD summaries of a shape from partial or local views, akin to other topological summaries already used as features in representation learning. After pretraining, the PH head can be discarded, leaving a backbone whose internal features are implicitly topology-aware yet free of PH computations at deployment.

\subsubsection{PH in Label Space and Output Calibration}

A fifth integration point lies in the \emph{output space} rather than the feature space. Even if the forward pass contains no explicit PH computation, PH can be applied post hoc to predicted masks, distance fields, or occupancy grids to support calibration, selection, and uncertainty estimation~\cite{clough2020topological,chen2019topological}.

In point cloud segmentation, per-point logits can be voxelized into binary masks for each class. Running PH on these masks reveals whether the predicted structures exhibit plausible connectivity patterns (e.g., buildings without spurious holes, roads without fragmented components). Instead of embedding PDs into the network, these signals can be used to select pseudo-labels in semi-supervised training: only predictions whose topology aligns with expected priors (or with ground truth PDs computed offline) are trusted, extending topology-constrained reconstruction and topological regularization ideas from images to point cloud outputs~\cite{clough2020topological,chen2019topological,jignasu2024stitch}. Similar ideas appear in topology-constrained reconstruction of implicit surfaces, where PH-based losses enforce, for example, that the reconstructed surface consists of a single connected component~\cite{jignasu2024stitch}.

PH also provides a structural measure of uncertainty. Predictions whose induced topology is highly unusual, such as façades with large unexpected voids or roofs with fragmented connectivity, can be flagged as low-confidence, prompting either human review or active learning queries~\cite{clough2020topological}. This topological notion of uncertainty complements more standard logit-, entropy-, or ensemble-based measures, and may be particularly valuable in safety-critical applications where structural validity is paramount~\cite{batra2019improved}.

\subsubsection{Learned PH on Weights and Feature Graphs}

Finally, PH can act on the \emph{network itself}, without ever touching the input point cloud. Neural persistence applies PH to the stratified graph of weights in a network, yielding a complexity measure that correlates with generalization and can guide tasks such as model selection, early stopping, or pruning~\cite{rieck2019neural}. More broadly, one can compute PH on intermediate feature graphs (e.g.~graphs of superpoints or region prototypes) and use the resulting invariants as regularizers, encouraging certain homology ranks or constraining the evolution of the feature topology during training~\cite{chen2019topological}.

Spectral constructions such as persistent Laplacians~\cite{memoli22persistent_laplacians} further blur the boundary between topological and geometric regularization. Their eigenvalues summarize PH together with geometric connectivity, and in principle can be used to shape the smoothness and multi-scale structure of learned representations. Viewed this way, PH becomes part of the architecture’s internal geometry, with the potential to influence capacity, sparsity, and robustness in ways that are orthogonal to explicit PD-based feature fusion.

\medskip
\paragraph{Summary} In summary, PH need not be confined to a parallel branch that computes diagrams, vectorizes them, and supplies an extra loss. Instead, PH can intervene at every stage of the pipeline: determining how point clouds are sampled and patched, reshaping neighborhood graphs and kernels, steering the training dynamics, structuring self-supervision, calibrating outputs, and regularizing the network’s own topology. For point cloud architectures, this broader view aligns naturally with recent calls to treat topological deep learning as a complement to geometric deep learning rather than a niche add-on~\cite{hensel21survey,papamarkou2024position}. It also opens a spectrum of practical compromises: PH can be used heavily during pre-processing or pretraining and then partially or fully removed at test time, allowing practitioners to benefit from topological inductive biases without incurring prohibitive runtime costs.

\section{Theoretical and Performance Implications}\label{sec:theorynperf}
\subsection{Theoretical Insights}
A key theoretical property of PH is \emph{stability}: small changes in input lead to small changes in the diagram. Specifically, there are metrics on diagrams like the \emph{Bottleneck distance} $d_B$ and the $p$-\emph{Wasserstein distance} $W_p$ which satisfy stability theorems~\cite{cohen2005stability,Cohen-Steiner2010}. Intuitively, $d_B(L,L')$ measures how far one must move points to match diagram $L$ to $L'$. Under mild assumptions, $d_B$ is bounded by the sup-norm of the difference between the underlying filtration functions~\cite{cohen2005stability}. Therefore, if the point cloud is slightly perturbed, its PD changes only slightly. In practice, this means PDs are robust to noise and geometric jitter~\cite{cohen2005stability}. While the stability theorem~\cite{cohen2005stability} ensures that the PD as a whole is robust to noise, the definition of \textit{diagonal gaps and subdiagrams} \cite{smith2021skeletonisation} provides a data-driven method to separate signal from noise by identifying the widest diagonal gaps. Further \cite{kudeshia2026learning} bridges these two concepts by proving that any feature identified as significant (with persistence $>\delta$) is guaranteed to persist in a perturbed dataset, provided the perturbation is small ($<\delta/2$).

However, since PDs are two-dimensional point clouds, it is crucial that any vectorization or embedding also preserves stability. PIs provably inherit 1-Wasserstein stability~\cite{adams2017persistence}, so that if two PDs are close, then their PIs are close in Euclidean space. This justifies using PIs in learning. Other constructions (landscapes, kernels) also satisfy similar stability properties~\cite{hofer2017deep,Cohen-Steiner2010}. In summary, topological features encode intrinsic shape properties that remain invariant under continuous transformations and small deformations, giving statistical robustness guarantees in 3D deep models.

PH also offers conceptual insights: by analyzing the evolution of features, one can infer object scale and connectivity patterns beyond what point coordinates alone reveal. For example, Grande and Schaub~\cite{grande2024non} recently showed that varying the underlying metric (non-isotropic PH) can extract additional orientation and scale information from 3D clouds. Filtration learning~\cite{nishikawa2023adaptive} is another emerging idea: instead of using a fixed radius function, a neural network learns the best filtration (e.g. an optimal height or density function) to highlight meaningful topological features for the task at hand. These advances enrich the theoretical toolbox for topology-based shape analysis. 
\begin{table*}[t]
\centering
\footnotesize
\caption{Comparison of complex construction time and number of simplices across varying point counts of a \textit{solid sphere and torus} point cloud (Fig. \ref{fig:commplex_vis}). For each sampling size, we report construction time and number of simplices.}
\label{tab:complex_construction}

\resizebox{\linewidth}{!}{%
\begin{tabular}{l|cc|cc|cc|cc|cc|cc}
\hline
\multirow{2}{*}{Construction} 
& \multicolumn{2}{c|}{1K pts} 
& \multicolumn{2}{c|}{10K pts} 
& \multicolumn{2}{c|}{50K pts} 
& \multicolumn{2}{c|}{100K pts} 
& \multicolumn{2}{c|}{500K pts} 
& \multicolumn{2}{c}{1M pts} \\
\cline{2-13}
& Time (s) & \#Simplices 
& Time (s) & \#Simplices
& Time (s) & \#Simplices
& Time (s) & \#Simplices
& Time (s) & \#Simplices
& Time (s) & \#Simplices \\
\hline
SSC      & 0.44118 & 147  & 0.92023 & 142  & 0.91501 & 147  & 0.92777 & 144  & 0.92155 & 148  & 0.91768 & 150 \\
Flood   & 0.08871 & 671   & 0.49493 & 4146  & 0.85859 & 8922  & 1.62552 & 17069 & 4.87386 & 36829 & 9.82121 & 55328 \\
Delaunay & 0.11202 & 8840  & 0.98786 & 54895 & 1.07122 & 68863 & 0.98627 & 68879 & 1.19278 & 68879 & 1.77467 & 68879 \\
Alpha    & 0.17922 & 4545  & 0.46643 & 25793 & 0.65020 & 31873 & 0.67728 & 31884 & 0.85199 & 31884 & 1.30659 & 31884 \\
Rips     & 0.01149 & 3155  & 0.41204 & 36985 & 1.44091 & 183788 & 2.41841 & 367500 & 10.30217 & 1844188 & 20.78028 & 3678235 \\
Witness  & 0.07758 & 556   & 0.04865 & 2078  & 0.16437 & 3947  & 0.21132 & 5338  & 0.87198 & 7393  & 1.65944 & 11312 \\
Cubical  & 0.00103 & 35937  & 0.00259 & 79507  & 0.01366 & 389017 & 0.03487 & 804357 & 0.22208 & 4019679 & 0.43830 & 7880599 \\
\hline
\end{tabular}
}
\end{table*}

\begin{figure}[t]
\centering
\includegraphics[width=\linewidth]{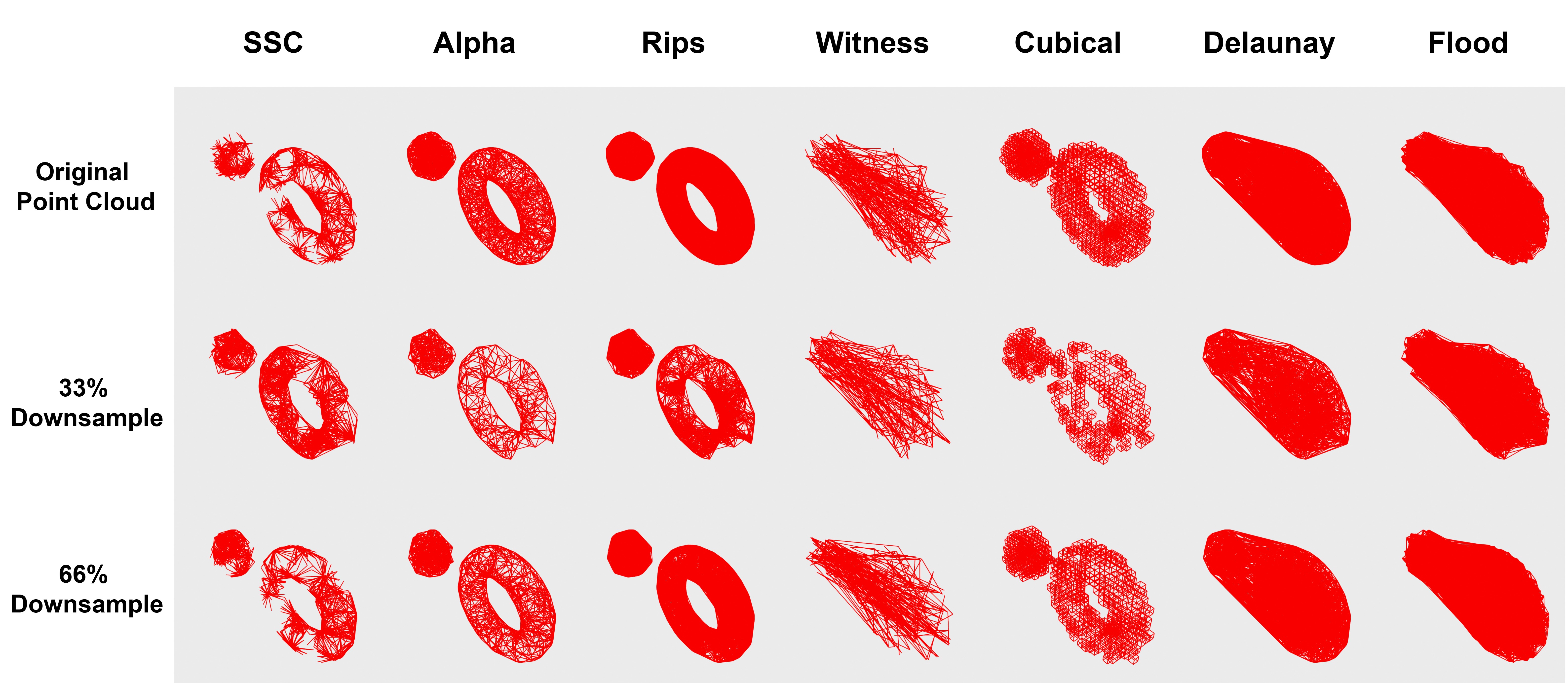}
\caption{Visualization of simplicial and cubical complexes constructed on a \textit{solid sphere and torus} point cloud under varying sampling densities. Rows correspond to 100K points, 33\% down-sampled, and 66\% down-sampled inputs, while columns show SSC \cite{kudeshia2024learning}, Alpha \cite{edelsbrunner2003shape}, Rips (VR) \cite{gromov1987hyperbolic}, Witness \cite{de2004topological}, Cubical \cite{kaczynski2004computing}, and Delaunay \cite{delaunay1934sphere} and Flood \cite{graf2025flood} complexes. The figure highlights qualitative differences in structural fidelity, sparsity, and geometric coherence across complex types as point density decreases.}
\label{fig:commplex_vis}
\end{figure}
\begin{figure}[h]
\centering
\includegraphics[width=8 cm]{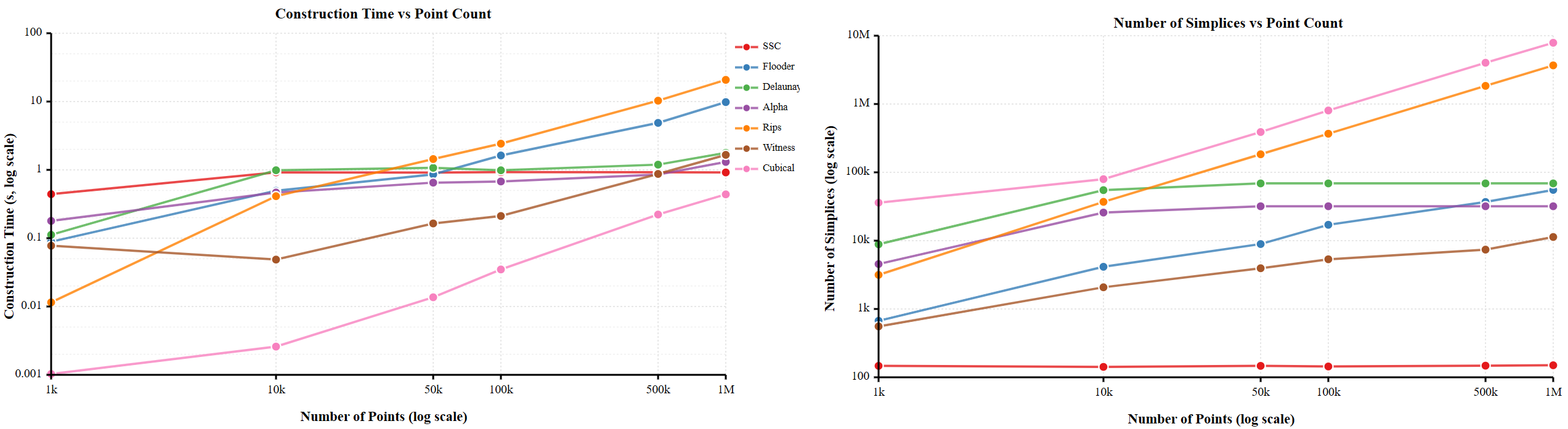}
\caption{Scaling behavior of complex construction methods across varying point cloud sizes (1k-1M points). Both plots use logarithmic scales on both axes. (a) Construction time vs point count: Cubical maintains the fastest construction (0.001-0.44s). SSC shows constant time (0.9s) across all scales. Rips shows exponential scaling (0.01→20.78s), becoming impractical beyond 500k points. (b) Number of simplices vs point count: SSC maintains constant minimal simplices (145-150). Cubical shows strong growth (36k→7.9M simplices). Delaunay and Alpha plateau beyond 50k points. Rips demonstrates explosive growth (3k→3.68M simplices). Logarithmic scales reveal distinct scaling regimes: constant (horizontal), sub-linear (gentle slopes), and exponential (steep slopes).}
\label{fig:scaling}
\end{figure}

\subsection{Performance Analysis}\label{sec:perf}
In practical 3D tasks, design choices in the PH pipeline significantly affect computational cost and empirical effectiveness. 

\paragraph{Complex construction and computational cost}  The computational burden of PH is sensitive to the number of simplices. For instance, a \v{C}ech complex has the desirable property of being homotopy-equivalent to the union of balls on the point cloud~\cite{dey2022computational}, but explicitly constructing it is prohibitively expensive in higher dimensions. Instead, many 3D works choose a Rips (VR) or alpha complex~\cite{zhou2022learning,Wong2021PHGCN,nishikawa2023adaptive,de2022ripsnet}. VR complexes are easy to compute from pairwise distances, but can grow combinatorially; alpha complexes restrict to Delaunay simplices, reducing size for well-sampled shapes. For volumetric data (e.g., occupancy grids or signed distance functions), cubical complexes are efficient since they align with voxel grids~\cite{gutierrez2012persistent,hu2024topology}. Recent work (Delaunay-Rips) even combines Delaunay and Rips filtration to balance completeness and speed~\cite{mishra2023stability}.

Fig.~\ref{fig:commplex_vis} shows different simplicial and cubical complexes constructed on a \textit{solid sphere and torus} point cloud to showcase the structural differences and sparsity of these complexes for three sample sizes of the original shapes. For further study on these complex constructions, Table \ref{tab:complex_construction} and Fig. \ref{fig:scaling} present a comparative analysis of the computational overhead associated with these simplicial complexes for a representative 3D shape \textit{solid sphere and torus}. In this quantitative assessment, we report the number of simplices and the computation time for different sample sizes ranging from $1,000$ to $1,000,000$ points. The cubical complex is the fastest construction overall, processing $1$ million points in just $0.438$ seconds, while complexes such as Witness and Alpha also maintain relatively low computation times. In terms of output complexity, the SSC is uniquely stable, consistently producing approximately $150$ simplices regardless of the input size. On the other hand, the Rips complex scales poorly, with its time and simplex count rising to $20.78$ seconds and over $3.6$ million at the largest scale. As shown in Fig. \ref{fig:scaling}, the flood complex shows faster construction for a smaller point size and moderate complexity at a higher sampling size. Moreover, unlike the Alpha or Delaunay complexes, its simplex count continues to grow steadily with the sample size, which makes it a denser representation than most of the other constructions. Therefore, the computational cost of simplicial complex construction grows rapidly with point cloud size and dimension, especially when such computations are performed per batch using tools like  GUDHI~\cite{maria14gudhi} or Ripser \cite{bauer2021ripser}. \textit{While precomputing topological features is feasible when the data or intermediate representations remain fixed (e.g., static raw point clouds), this becomes infeasible in end-to-end training pipelines where feature clouds evolve at each iteration due to parameter updates.}

To mitigate this cost, one can potentially consider several strategies:
\begin{itemize}
    \item \textbf{Subsampling.} Reduce the number of points before complex construction (e.g., applying Farthest Point Sampling (FPS) using \texttt{gudhi.subsampling} or similar functions, thereby controlling combinatorial explosion in the complex construction.
    \item \textbf{Faster or approximate representations.} Replace full PDs with computationally cheaper summaries such as PLs or PIs. Alternatively, use faster libraries like Ripser instead of GUDHI.
    \item \textbf{Caching and asynchronous evaluation.} If batches repeat or intermediate feature representations stabilize, cache and reuse computed topological signatures (e.g., with a \texttt{PDCache}). Topological computations can also be offloaded asynchronously to avoid blocking the training loop.
\end{itemize}

Ultimately, the computational cost stems from simplex enumeration and boundary matrix reduction. When topological descriptors evolve dynamically with model training, real-time computation is often necessary; thus, approximations, backend selection, and memory-aware batching become essential for scalability.

\begin{table}[t]
\centering
\footnotesize
\caption{Topological stability measured by the 2-Wasserstein distance 
$W_2^{(k)} = W_2(\mathcal{D}_k(P), \mathcal{D}_k(P_\sigma))$ between PDs of the original point cloud $P$ and its Gaussian-noised version $P_\sigma$ for different noise levels $\sigma$.}
\label{tab:wd_100k_sigma_split}

\resizebox{\linewidth}{!}{%
\begin{tabular}{l|cc|cc|cc|cc}
\hline
\multirow{2}{*}{Complex} 
& \multicolumn{2}{c|}{$\sigma=0.010$} 
& \multicolumn{2}{c|}{$\sigma=0.020$} 
& \multicolumn{2}{c|}{$\sigma=0.050$} 
& \multicolumn{2}{c}{$\sigma=0.100$} \\
\cline{2-9}
& $W_2^{(0)}$ & $W_2^{(1)}$
& $W_2^{(0)}$ & $W_2^{(1)}$
& $W_2^{(0)}$ & $W_2^{(1)}$
& $W_2^{(0)}$ & $W_2^{(1)}$ \\
\hline
Flood   & 0.44 & 0.30 & 0.67 & 0.47 & 0.69 & 0.63 & 0.59 & 0.72 \\
Alpha     & 3.11 & 0.88 & 4.77 & 1.57 & 6.57 & 2.29 & 7.86 & 2.63 \\
Cubical   & 0.24 & 0.00 & 0.33 & 0.00 & 0.56 & 0.00 & 0.71 & 0.00 \\
Rips (VR) & 0.67 & 0.37 & 0.45 & 0.39 & 0.57 & 1.22 & 2.66 & 2.06 \\
Witness   & 0.00 & 0.11 & 0.00 & 0.17 & 0.00 & 0.20 & 0.00 & 0.20 \\
\hline
\end{tabular}%
}
\end{table}

\begin{table}[t]
\centering
\footnotesize
\caption{Topological stability measured by the 2-Wasserstein distance 
$W_2^{(k)} = W_2(\mathcal{D}_k(P), \mathcal{D}_k(P_s))$ between PDs of the original point cloud $P$ (1M points) and its downsampled versions $P_s$ for different downsampling factors.}
\label{tab:wd_downsample}
\resizebox{\linewidth}{!}{%
\begin{tabular}{l|cc|cc|cc|cc|cc}
\hline
\multirow{2}{*}{Complex}
& \multicolumn{2}{c|}{500k} 
& \multicolumn{2}{c|}{100k} 
& \multicolumn{2}{c|}{50k} 
& \multicolumn{2}{c|}{10k} 
& \multicolumn{2}{c}{1k} \\
\cline{2-11}
& $W_2^{(0)}$ & $W_2^{(1)}$
& $W_2^{(0)}$ & $W_2^{(1)}$
& $W_2^{(0)}$ & $W_2^{(1)}$
& $W_2^{(0)}$ & $W_2^{(1)}$
& $W_2^{(0)}$ & $W_2^{(1)}$ \\
\hline
Flood   & 0.046 & 0.119 & 0.034 & 0.123 & 0.050 & 0.128 & 0.086 & 0.152 & 0.714 & 0.684 \\
Alpha     & 0.000 & 0.000 & 0.000 & 0.000 & 0.000 & 0.000 & 0.832 & 0.408 & 3.606 & 1.703 \\
Cubical   & 0.000 & 0.000 & 0.000 & 0.000 & 0.000 & 0.000 & 0.134 & 0.000 & 0.523 & 0.000 \\
Rips (VR) & 0.433 & 0.422 & 0.221 & 0.418 & 0.313 & 0.380 & 0.211 & 0.411 & 0.730 & 0.293 \\
Witness   & 0.140 & 0.068 & 0.089 & 0.045 & 0.085 & 0.066 & 0.059 & 0.051 & 0.085 & 0.100 \\
\hline
\end{tabular}%
}
\end{table}

\begin{figure}[t]
     \centering
     \begin{subfigure}[b]{0.48\textwidth}
         \centering
         \includegraphics[width=\linewidth]{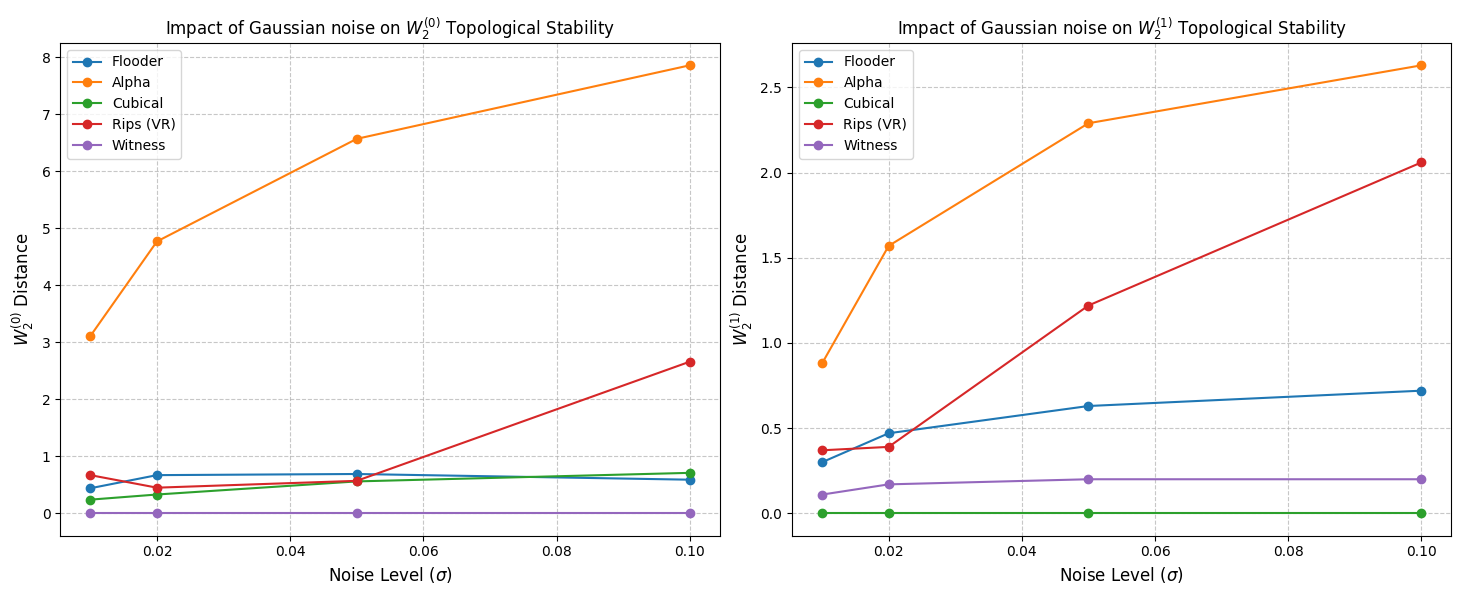}
         \caption{Stability vs. Noise Level $\sigma$}
         \label{fig:stabilityNoise}
     \end{subfigure}
     \hfill
     \begin{subfigure}[b]{0.48\textwidth}
         \centering
         \includegraphics[width=\linewidth]{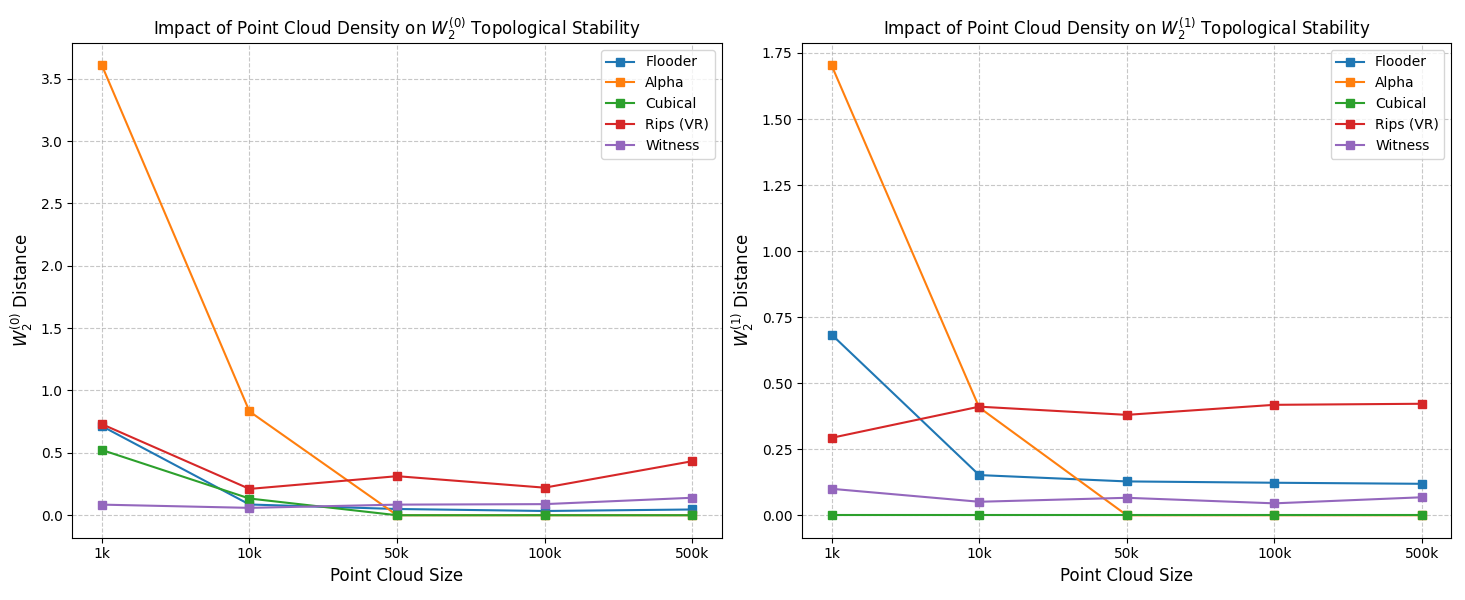}
         \caption{Stability vs. Downsampling Factor}
         \label{fig:stabilitySize}
     \end{subfigure}
     
     \caption{Topological stability measured by the 2-Wasserstein distance $W_2^{(k)} = W_2(\mathcal{D}_k(P), \mathcal{D}_k(P_s))$ for both Gaussian noise (left) and downsampling (right).}
     \label{fig:stabilityComparison}
\end{figure}

\paragraph{Feature vectorization and efficiency} Translating a diagram into a vector also has cost implications. A fine-grained grid for PIs (e.g., 100$\times$100) yields high-dimensional input vectors. In large scenes, PDs may contain thousands of points, making naive binning expensive. Nevertheless, vectorization remains faster than, say, computing complex kernels for each training batch~\cite{hofer2017deep}. To reduce dimensionality, many 3D PH methods focus on only low-dimensional homology: since shapes live in $\mathbb{R}^3$, they often ignore $H_k$ for $k>2$. For instance, Hu \etal~\cite{hu2024topology} generate shapes by tracking only 1D holes, and Beksi \etal~\cite{beksi20163d} used only $H_0$ features for segmentation. This selective filtration can dramatically lower computational load. Once vectors are extracted, they are typically appended to standard point features (e.g., edge lengths, normals) and fed into architectures like DGCNN~\cite{wang2019dynamic} or DeepGCN~\cite{li2019deepgcns}. 

\paragraph{Accuracy} Empirically, topology-augmented models often outperform pure-geometry baselines. For example, PHGCN~\cite{Wong2021PHGCN} achieves significantly higher segmentation accuracy on ShapeNet and PartNet compared to DGCNN or ResGCN alone. This suggests topological features capture fine-scale structure (thin parts, loops) that geometry misses. Likewise, Zhou \etal~\cite{zhou2022learning} report that adding PI features yields gains in classification and detection. Turkes \etal~\cite{Turkes2022} quantitatively showed that PH-based networks better recover the number of holes in noisy 3D shapes than conventional nets. These studies indicate that PH provides valuable invariants for shape learning. These results motivate a systematic empirical study in Section~\ref{sec:experiments}, where we augment three representative point cloud classification and part segmentation backbones with PI-, PL-, and PD- features to analyze their relative impact on performance.

\paragraph{Topological Stability}Stability experiments also confirm robustness: Zhou \etal \cite{zhou2022learning} observed that their PI-based classifier’s accuracy degrades only slightly under moderate Gaussian noise. Similarly, TopoSeg~\cite{Liu2022} reports stable part segmentation under small perturbations. In short, persistent topological descriptors provide a noise-tolerant complement to Euclidean features. To study the topological stability of various simplicial complexes for different Gaussian noise and varying sampling densities, we vary Gaussian noise levels from 0.010 to 0.100 and point cloud sampling sizes from 1k to the original size, which is 1M points, and report 2-Wasserstein distances $W_2^{(0)}$ and $W_2^{(1)}$ between PDs of the original point cloud $P$ (1M points) and its downsampled versions $P_s$, see Tables \ref{tab:wd_100k_sigma_split}-\ref{tab:wd_downsample}. As shown in Fig. \ref{fig:stabilityNoise}, while the 2-Wasserstein distances ($W_2^{(0)}$ and $W_2^{(1)}$) generally increase with higher noise levels ($\sigma$) in the point cloud, different filtration exhibit varying degrees of sensitivity. For instance, Cubical, Witness, and flood filtration maintain a remarkably low and stable distance across all noise levels, indicating that they are less prone to the noise caused by sensor jitter. The impact of point cloud density, illustrated in Fig. \ref{fig:stabilitySize}, further highlights the scale-invariant nature of these signatures. For most filtration methods, the 2-Wasserstein distances ($W_2^{(0)}$ and $W_2^{(1)}$) drop sharply and plateau as the point cloud size increases, suggesting that the captured topological features remain consistent despite sparse sampling. This behavior confirms that topological signatures provide a reliable, noise-tolerant representation, allowing deep learning backbones to focus on the underlying global structure of a 3D object rather than minor surface irregularities or sampling gaps.

\begin{figure*}[t]
\centering
\includegraphics[width=16 cm]{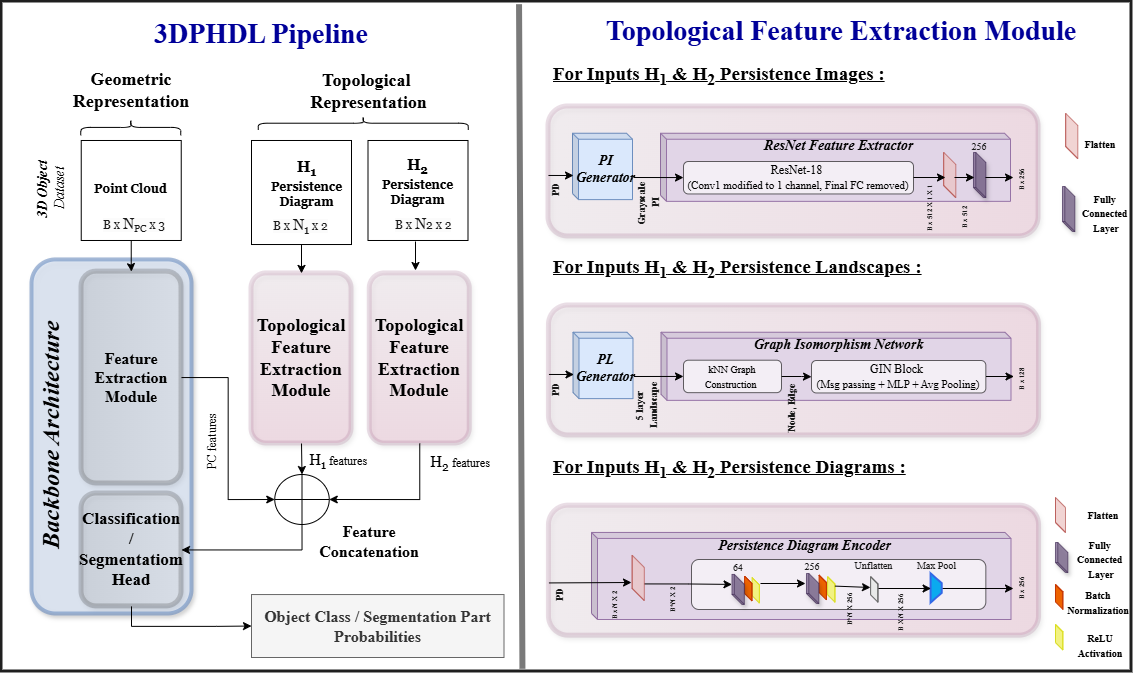}
\caption{Schematic diagram of modified backbone architectures used for the 3D object classification or part segmentation experiments.}
\label{fig:arch}
\end{figure*}

\paragraph{Scalability challenges} The flip side of utilizing PH is the computational overhead. PH has worst-case $O(n^3)$ complexity, and real 3D scenes can have thousands of points, yielding millions of simplices in a Rips complex. Even though practical heuristics (subsampling, dimension truncation) are used, training on full-scale point clouds with PH vectors is still slower. Vector dimensionality also grows with the number of features; very fine binning leads to large PI vectors that strain memory. Future work must address scaling PH: possible solutions include multi-scale persistence, graph sparsification, or GPU-accelerated PH~\cite{Bauer2014,mishra2023stability}. Another approach is to limit homology dimensions to task-relevant ones (e.g. ignore $H_2$ for thin structures) as already done in some works.

\paragraph{Topological loss and supervision} Incorporating topology into the loss requires differentiable surrogates of PD distances. Existing methods use MSE on PIs~\cite{zhou2022learning} or approximate Wasserstein distances~\cite{Wong2021PHGCN,Liu2022}. However, the best design for such losses remains an open question. Moreover, fully leveraging topology in training may need datasets with ground-truth topological labels (e.g. objects annotated by number of holes). Most current experiments rely on existing benchmarks (ModelNet~\cite{Chang2015ShapeNet}, ShapeNet~\cite{Chang2015ShapeNet}, PartNet~\cite{Mo2019}) or synthetically generate topology. Creating benchmark datasets specifically for topological tasks (e.g. hole detection) would accelerate research. 

\begin{table*}[t]
\centering
\footnotesize
\caption{Effect of topological vectorization choices across point-cloud backbones on ModelNet40~\cite{wu20153d} classification.
We compare Rips-complex-based persistence vectorization (PI, landscapes, PD) against non-topological baselines for PointNet~\cite{pointnet}, DGCNN ~\cite{wang2019dynamic}, and  PointTransformer \cite{zhao2021point}, reporting accuracy, robustness metrics, and model complexity. $T_{e}$ denotes average training time per epoch; $T_{inf}$ denotes inference time.}
\label{tab:exp_cls}
\resizebox{\linewidth}{!}{%
\begin{tabular}{ccccccccccc}
\hline
Backbone & Complex & Vectorization  & Accuracy & mAcc & Precision & Recall & F1 & Params (M) & $T_{e}$(s) & $T_{inf}$(s)\\
\hline

\multirow{4}{*}{PointNet~\cite{pointnet}}
& --   & --        & 89.2 & 86.1 & 86.3 & 85.9 & 86.1 & 3.4  & 3.5 & 1.9 \\
& Rips & PI        & 90.4 & 86.9 & 87.7 & 86.2 & 86.9 & 26.3 & 1361.5 & 273.6  \\
& Rips & Landscape & 90.7 & 87.3 & 87.4 & 86.6 & 87.0 & 3.6  & 2672.4 & 428.1 \\
& Rips & PD        & 89.2 & 86.8 & 86.9 & 86.0 & 86.4 & 3.7  & 10.7 & 1.9\\

\hline

\multirow{4}{*}{DGCNN~\cite{wang2019dynamic}}
& --   & --        & 92.9 & 90.2 & 90.8 & 90.1 & 90.4 & 1.8  & 15.3 & 15.7  \\
& Rips & PI        & 93.5 & 90.9 & 91.1 & 90.6 & 90.8 & 24.6 & 1368.6 & 282.4  \\
& Rips & Landscape & 93.9 & 92.1 & 91.3 & 90.8 & 91.1 & 1.9  & 4524.5 & 423.6 \\
& Rips & PD        & 93.2 & 91.7 & 90.9 & 90.3 & 90.6 & 2.1  & 76.4 & 17.1  \\

\hline

\multirow{4}{*}{PointTransformer~\cite{zhao2021point}}
& --   & --        & 93.4 & 90.2 & 88.4 & 89.8 & 89.1 & 3.9 & 291.8 & 72.9 \\
& Rips & PI        & 94.6 & 91.2 & 89.9 & 90.5 & 90.3 & 26.6 & 1488.0 & 358.3 \\
& Rips & Landscape & 94.8 & 90.9 & 90.0 & 90.5 & 90.1 & 4.0 & 2642.0 & 542.1\\
& Rips & PD        & 93.5 & 90.4 & 90.1 & 89.9 & 90.5 & 4.1 & 292.9 & 50.8 \\

\hline
\end{tabular}
}
\normalsize
\end{table*}
\begin{figure*}[h!t]
\centering
\hspace*{-1cm}
\includegraphics[width=16 cm]{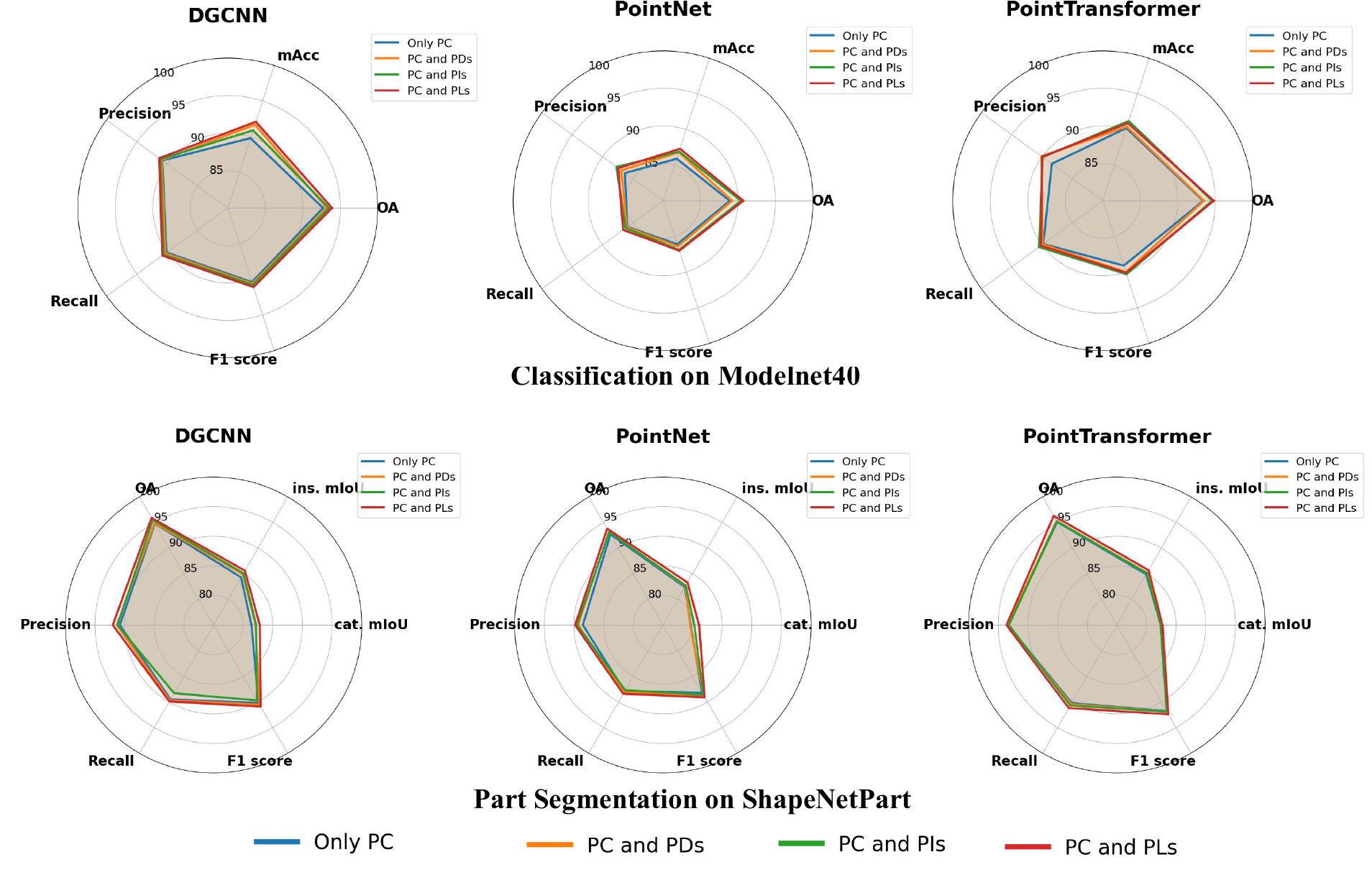}
\caption{ Performance comparison of vectorization methods for each backbone model for classification (left) and part-segmentation (right). Incorporating persistence-based vectorizations (PDs, PIs, PLs) yields consistent gains in accuracy and balance metrics (Precision/Recall/F1), demonstrating that topological summaries capture discriminative shape characteristics not fully represented by point geometry alone. }
\label{fig:radar}
\end{figure*}

\section{3DPHDL Pipeline: An Empirical Study}\label{sec:experiments}

In this section, we provide an empirical study of the 3DPHDL pipeline across two standard downstream tasks: 3D object classification and part segmentation. In all experiments, the PH stack comprising complex construction, filtration, and persistence computation is fixed to a computationally tractable configuration, while we vary only a small set of design choices at the vectorization and backbone stages. This restricted setup is sufficient to demonstrate that even a single, practical instantiation of the 3DPHDL pipeline can yield competitive performance and informative behavior on representative benchmarks, without claiming a comprehensive search over all possible design combinations.

To incorporate topological information, we construct simplicial complexes on each point cloud and apply a VR filtration to compute PH. This process yields PDs, which summarize the multi-scale topological features (e.g., connected components, loops, and voids) of the underlying finite metric spaces. These PDs are further vectorized during model training and inference to serve as inputs to the learning model.

For these experiments, we employ the VR complex, as it stands as the standard methodology for generating filtrations from geometric datasets due to its balance of computational tractability and theoretical robustness \cite{de2004topological, bauer2021ripser}. VR complex is a clique complex and hence determined entirely by its 1-skeleton, making it convenient to implement and highly efficient for integration into deep learning pipelines. Furthermore, the VR complex provides a stable topological representation that is guaranteed to recover the homotopy type of the underlying manifold under sufficient sampling density \cite{hausmann1995vietoris}, ensuring that the features extracted by our model accurately reflect the data's global structure even in the presence of noise \cite{chazal2014persistence}

For 3D object classification and part segmentation, we use three different point cloud learning backbone architectures: PointNet~\cite{pointnet}, DGCNN ~\cite{wang2019dynamic}, and  PointTransformer \cite{zhao2021point}. These architectures are selected to represent the dominant paradigms in point cloud processing. PointNet~\cite{pointnet} serves as the foundational MLP-based baseline for global feature aggregation. DGCNN ~\cite{wang2019dynamic} utilizes dynamic graph convolutions to capture local geometric structures. Point Transformer \cite{zhao2021point} leverages the self-attention mechanism to model complex long-range dependencies within the 3D data. Together, they provide a comprehensive benchmark across point-based, graph-based, and transformer-based learning strategies.

To integrate topological information in the backbone architectures, we introduce additional branches with different topological vectorization inputs, such as $H_{1}$ and $H_{2}$ PIs, $H_{1}$ and $H_{2}$ PLs, and $H_{1}$ and $H_{2}$ PDs to extract features from each input modality which are concatenated with the point cloud features and further fed to classification or segmentation head. As shown in Fig.~\ref{fig:arch}, for each vectorized topological input modality, we use a different branch architecture. For PIs, we use a feature-extraction layer based on ResNet-18 \cite{torchvision2016} for each of the $H_{1}$ and $H_{2}$ PIs. For PLs, we use a layer of graph isomorphism network (GIN) to extract features from each $H_{1}$ and $H_{2}$ PL individually. To learn features from $H_{1}$ and $H_{2}$ PDs, we use MLP-based feature extractor layers for each diagram.

All the architectures are developed in Python 3.11 and utilize PyTorch 2.0 for artificial neural network layers. For both tasks: classification and part segmentation, all models are trained and tested on an AMD EPYC 9454 (Zen 4) CPU with an NVIDIA H100 SXM5 GPU.


\subsection{3D Object Classification (ModelNet40)}

We perform classification on the ModelNet40 dataset~\cite{wu20153d}, consisting of 12,311 point cloud models across 40 object categories. For evaluation metrics, in Table \ref{tab:exp_cls}, we report overall accuracy, average class-wise accuracy (mAcc), precision, recall, F1-score, model parameters (Params), average training time per epoch $T_{e}$, and inference time $T_{inf}$. 

 The empirical results shown in Table \ref{tab:exp_cls} demonstrate that augmenting standard point cloud backbone architectures with additional topological information consistently yields superior 3D object classification performance compared to baseline networks. Integration of PH features through various vectorization, specifically, the PL vectorization, emerges as a highly efficient and effective choice as it shows measurable gains in Accuracy, Precision, Recall, and F1-score while maintaining the model parameter count comparable to original backbone architectures. Furthermore, as shown in Fig.~\ref{fig:radar}, the simultaneous improvement in the Precision as well as in the Recall metrics across most topological variants suggests that topological features enhance the model's structural awareness and aid in effectively distinguishing between classes with similar global appearances but distinct topological signatures.




\begin{table*}[t]
\centering
\footnotesize
\caption{Part-segmentation performance on ShapeNetPart~\cite{yi2016scalable} using different topological vectorization across backbones. 
We compare Rips-complex-based persistence vectorization (PI, Landscape, PD) against non-topological baselines for PointNet~\cite{pointnet}, DGCNN ~\cite{wang2019dynamic}, and  PointTransformer \cite{zhao2021point}, reporting accuracy, mean IoUs, robustness metrics, and model complexity. $T_{e}$ denotes average training time per epoch; $T_{inf}$ denotes inference time.}
\label{tab:exp_seg}
\resizebox{\linewidth}{!}{%
\begin{tabular}{cccccccccccc}
\hline
 Backbone & Complex & Vectorization  & cat. mIoU & ins. mIoU & Accuracy & Precision & Recall & F1 & Params (M) & $T_{e}$(s) & $T_{inf}$(s) \\
\hline

 \multirow{4}{*}{PointNet~\cite{pointnet}} 
& -- & --        & 79.7 & 82.4 & 92.7 & 88.5 & 87.9 & 88.2 & 8.3  & 38.9 & 18.1 \\
  & Rips & PI        & 80.3 & 82.7 & 93.1 & 89.5 & 87.7 & 88.6 & 31.0 & 1033.7 & 264.1 \\
  & Rips & Landscape & 81.1 & 83.3 & 93.8 & 89.8 & 88.4 & 89.1 & 8.4  & 2170.4 & 519.4  \\
  & Rips & PD        & 79.6 & 82.6 & 93.2 & 89.3 & 88.1 & 88.7 & 8.5  & 40.8	& 57.7 \\

\hline

 \multirow{4}{*}{DGCNN~\cite{wang2019dynamic}} 
& -- & --        & 81.4 & 84.3 & 94.8 & 90.8 & 89.5 & 90.2 & 1.4  &  84.3 & 47.1\\
  & Rips & PI        & 82.2 & 85.1 & 95.5 & 91.2 & 88.3 & 89.7 & 24.1 &  1089.5 & 283.9\\
  & Rips & Landscape & 82.8 & 85.6 & 95.9 & 92.0 & 89.9 & 90.9 & 1.5  & 2193.2 & 520.9 \\
  & Rips & PD        & 82.1 & 84.9 & 94.9 & 91.3 & 89.7 & 90.5 & 1.6  & 88.1 & 153.3 \\

\hline

 \multirow{4}{*}{PointTransformer~\cite{zhao2021point}} 
& -- & --        & 82.4 & 84.9 & 95.1 & 93.4 & 90.2 & 91.8 & 5.3  & 172.4 & 161.9 \\
  & Rips & PI        & 82.4 & 85.1 & 95.2 & 93.2 & 90.7 & 91.9 & 28.0 & 1164.7 & 275.3 \\
  & Rips & Landscape & 82.7 & 85.7 & 96.3 & 93.6 & 91.2 & 92.4 & 5.4  & 2276.0 & 595.8 \\
  & Rips & PD        & 82.5 & 85.2 & 95.3 & 93.4 & 90.4 & 91.9 & 5.6  & 178.3 & 226.9 \\

\hline
\end{tabular}
}
\normalsize
\end{table*}
\begin{figure*}[h!t]
\centering
\includegraphics[width=16 cm]{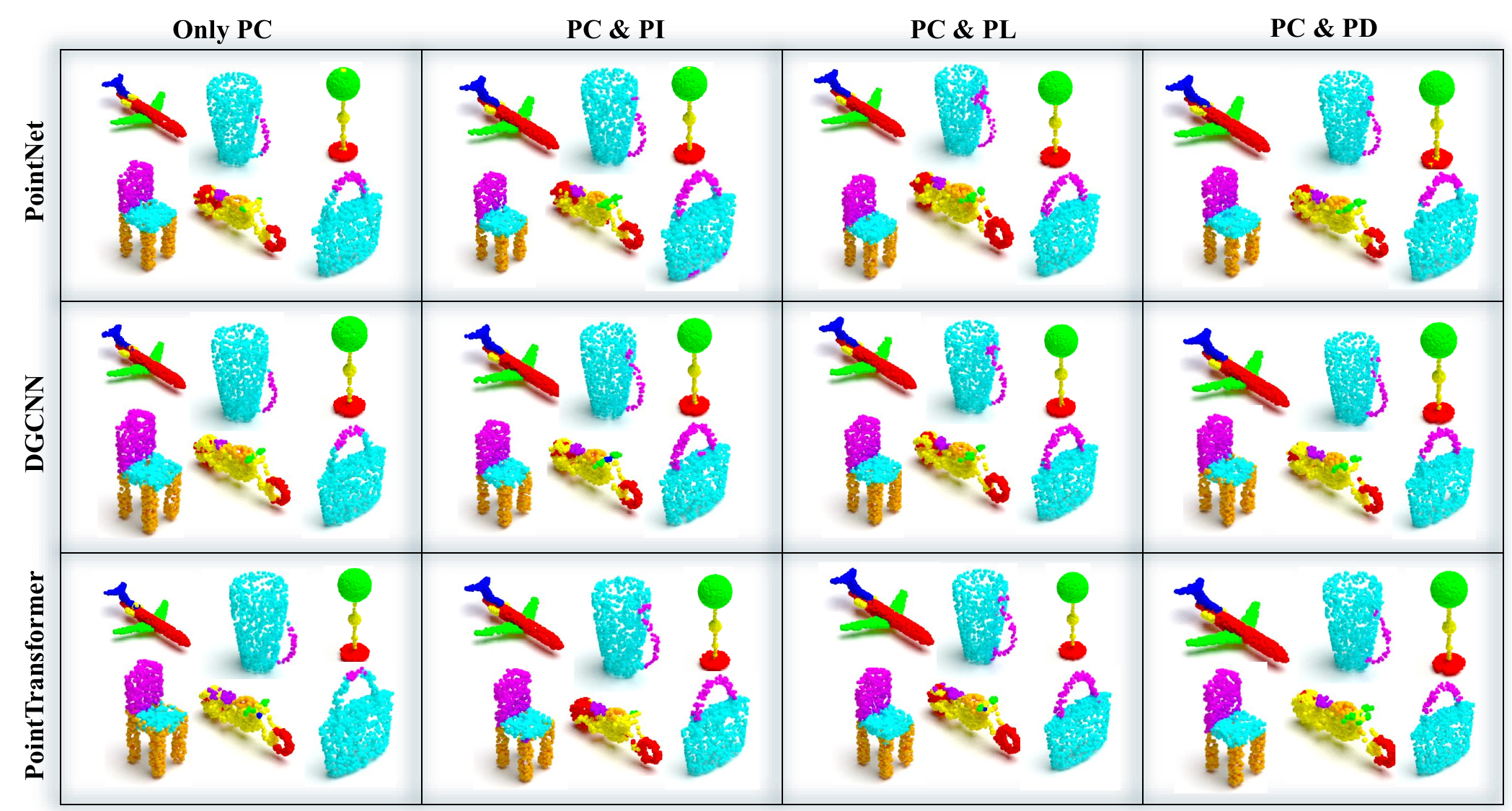}
\caption{Qualitative visualization of learned point cloud representations for part segmentation on ShapenetPart across different backbone architectures. Each row presents the same six object categories (airplane, mug, lamp, chair, motorbike, bag) as processed by PointNet (top), DGCNN (middle), and PointTransformer (bottom). Each column corresponds to a different backbone architecture with different input modalities. \textbf{Only PC} refers to the backbone architecture having a single branch with point cloud input only. 
\textbf{PC \& PI}, \textbf{PC \& PL}, and \textbf{PC \& PD} refer to three-branch architectures where one branch processes the point cloud, and the other two branches process the $H_{1}$ and $H_{2}$ persistent homology features, vectorized using persistence images (PI), persistence landscapes (PL), and persistence diagrams (PD), respectively.}

\label{fig:segmentationViz}
\end{figure*}

\subsection{Part Segmentation (ShapeNetPart)}
We use ShapeNetPart~\cite{yi2016scalable}, containing 16,880 models across 16 object categories and 50 parts. For evaluation metrics, we report average category Intersection-over-Union (cat. mIoU), average instance Intersection-over-Union (ins. mIoU), accuracy, precision, recall, F1-score, model parameters (Params), average training time per epoch $T_{e}$, and inference time $T_{inf}$.

As shown in Table \ref{tab:exp_seg} and Fig.~\ref{fig:radar}, the integration of topological information into each backbone architecture leads to consistent improvements in category mIoU, instance mIoU, accuracy, precision, recall, and F1-score compared to using point cloud input alone. Among the different topological representations, PL yields the best overall performance, followed by PI, which is having substantially larger number of model parameters. Additionally, the consistent improvement in F1-score with the simultaneous gain in Precision and Recall provided by the PL-based models across different architectures demonstrates a superior ability to capture the complex geometric and topological features necessary for precise part boundaries and helps the models better distinguish between fine-grained part segments. Therefore, an additional PL input achieves the best overall performance with a relatively small increase in model size, demonstrating that topological features can significantly enhance segmentation performance even under additional computational overhead due to the on-the-fly vectorization of topological information.

Fig.~\ref{fig:segmentationViz} presents a part segmentation visualization of six different ShapeNet objects using each backbone architecture with only point cloud inputs as well as additional topological input modalities. We observe that for each backbone architecture, models with additional topological inputs, especially the PLs and PIs, aid in understanding the topology of the shape and hence show significant improvement in the segmentation of a few parts, such as the handle of a bag and a mug, or the wheel of a motorbike. Similarly, a noticeably improved segmentation of the shade of the lamp in the PointNet backbone and the seat and legs of the chair in pointTransformer architectures can be observed when these models are incorporated with the topological inputs. Therefore, these results demonstrate the effectiveness of topological information in point-cloud understanding, and the topologically integrated models show improved segmentation results in comparison to backbone networks solely relying on geometric point cloud input.

These ablations explore the influence of PH at different levels. For classification and segmentation, PH improves accuracy and part consistency. 
Although a higher computational cost is incurred due to the complex construction, filtration, and vectorization of topological information, the integration of topological information still provides a favorable trade-off between performance and complexity and hence provides actionable guidelines for integrating topological reasoning into 3D deep learning pipelines.

\section{Conclusion and Future Directions}\label{sec:conclusion}

\paragraph{Summary}Persistent homology has emerged as a distinctive complement to geometric and spectral tools for 3D shape analysis. Its stability theorems guarantee robustness, while its invariants summarize multi-scale connectivity, separation, and void structure that are not directly visible to purely local geometric descriptors. This paper reframes \textbf{3DPHDL} not as a single recipe but as a design space. The central PHML pipeline (complex construction, filtration, PH computation, vectorization, neural architectures, and prediction) organizes the core choices, while the six injection points (IP1-IP6) show how PH can intervene at every level, e.g.,  in sampling and patching, in neighborhood graphs and kernels, in training dynamics, in self-supervision and augmentation, in label-space calibration, and in the network's own weight and feature topology. Recent advances such as learned filtrations for point clouds~\cite{nishikawa2023adaptive}, non-isotropic persistent homology that exploits metric dependence~\cite{grande2024non}, and topology-aware generative models for 3D content~\cite{hu2024topology} demonstrate that PH can be made task-adaptive, anisotropic, and generative rather than static and hand-crafted. At the same time, PH computation remains a bottleneck at scale, most empirical studies still focus on moderate-size or synthetic data, and there is a lack of benchmarks with explicit topological annotations, all of which limit systematic evaluation of 3DPHDL design choices~\cite{Bauer2014,mishra2023stability}.

\paragraph{Conclusions}As surveyed in this work, PH features enrich deep learning on point clouds, improving tasks that depend on global structure, such as connectivity-aware segmentation, and topology-sensitive classification, especially under noise and sampling artifacts. While the proposed framework formalizes a six-dimensional PH design space, the present empirical study instantiates a controlled slice of this space by fixing the PH representation to PD, the complex construction to the VR complex, and the filtration to a distance-based scheme. Within this setting, we systematically vary three principal axes: backbone architecture, PH vectorization derived from PDs, and prediction task. We focused on two fundamental geometric understanding tasks-classification and segmentation which evaluate both global and part-level structural reasoning. Other task categories within the design space, such as regression (e.g., geometric attribute prediction) and reconstruction (e.g., topology-aware shape generation), remain open for systematic study. The remaining axes of the design space including alternative PH representations, density-driven or learned filtrations, and different complex constructions are explicitly defined within our framework and provide structured directions for future exploration.

\paragraph{Future Outlook}Looking ahead, several deeper directions emerge. First, there is a clear opportunity to move from \enquote{PH as a side branch} to architectures where topology and geometry co-evolve: learned filtrations and non-isotropic metrics can be trained jointly with geometric backbones so that the network discovers which topological events, in which regions and at which scales, are discriminative for a task~\cite{nishikawa2023adaptive,grande2024non}. Second, approximate or surrogate PH layers, ranging from neural estimators of persistence (e.g.\ neural persistence and related complexity measures) to spectral operators such as persistent Laplacians, could make topology-aware training compatible with large-scale point clouds, enabling IP2-IP3-IP4 style integration (topology-aware neighborhoods, training dynamics, and self-supervision) without explicit diagram computation at every iteration~\cite{rieck2019neural,memoli22persistent_laplacians}. Third, PH applied to outputs and weights (IP5-IP6) suggests a research program on topology-aware reliability: using PH to calibrate label-space structure, to define curriculum and hard-example policies, and to monitor network complexity and internal failure modes in safety-critical 3D perception~\cite{clough2020topological,chen2019topological,batra2019improved}. Fourth, an important open question concerns representation-level topology in large pretrained point-cloud models~\cite{saito2025point, pang2023masked, yu2022point}. It remains unclear to what extent their learned latent spaces implicitly encode persistent topological structure, and whether PH computed on intermediate activations can reveal stable geometric invariants. Conversely, PH-guided objectives could be used to explicitly encourage foundation models to preserve task-relevant topological signals during pretraining. Understanding the interplay between self-supervised representation learning and persistent topology represents a promising direction toward topology-aware foundation models for 3D data.

\paragraph{Final Remarks}Ultimately, \textbf{3DPHDL} invites a shift in how topology is viewed within geometric deep learning; not as a niche regularizer or visualization tool, but as a structural inductive bias that permeates the data pipeline, the architecture, and the optimization scheme. Realizing this vision will require scalable PH backends (e.g.\ distributed, streaming, and GPU-accelerated algorithms), benchmarks with controlled topological variation, and theory that links PH-based complexity measures to generalization and robustness of 3D networks~\cite{Bauer2014,mishra2023stability,rieck2019neural}. If these ingredients come together, persistent homology can become part of the internal geometry of 3D deep learning systems, enabling models that perceive and reason about global shape structure as naturally as they currently exploit local metric cues.



\bibliographystyle{elsarticle-num}
\bibliography{ph_topology}
\end{document}